\documentclass[journal]{IEEEtran}

\usepackage{xcolor}
\usepackage{graphicx}
\usepackage{float}
\usepackage{amsmath}
\usepackage{xltabular}
\usepackage{booktabs}
\usepackage{pdflscape}
\usepackage{adjustbox}
\usepackage{multirow}
\usepackage{comment} 
\usepackage{makecell}
\usepackage{array}
\usepackage{booktabs}
\usepackage{hyperref}
\usepackage{booktabs} % For formal tables
\usepackage{longtable} % For tables that span multiple pages
\usepackage[numbers,sort&compress]{natbib}
\usepackage{listings}
\ifCLASSINFOpdf
\else
\fi
\begin{document}
\title{EMT: A Visual Multi-Task Benchmark Dataset for Autonomous Driving in the Arab Gulf Region}

\author{Nadya Abdel Madjid$^{1,4*}$, Murad Mebrahtu$^{1,4*}$, Abdulrahman Ahmad$^{1,4}$, Abdelmoamen Nasser$^{1,4}$,  Bilal Hassan$^{2}$, Naoufel Werghi$^{1}$,  Jorge Dias$^{3,4}$, and Majid Khonji$^{1,4}$% 
\thanks{This work was supported by Khalifa University of Science and Technology under Award No. RIG-2023-117. Corresponding authors: nadya.abdelmadjid@gmail.com, murad.mebrahtu@ku.ac.ae, majid.khonji@ku.ac.ae}% 
\thanks{$^{*}$ Equal contribution}
\thanks{$^{1}$ Computer Science, Khalifa University, Abu Dhabi, UAE}%
\thanks{$^{2}$ Computer Science, New York University, Abu Dhabi, UAE}%
\thanks{$^{3}$ Computer and Information Engineering, Khalifa University, Abu Dhabi, UAE}%
\thanks{$^{4}$ KUCARS-KU Center for Autonomous Robotic Systems, Khalifa University, Abu Dhabi, UAE}%
}


\markboth{Journal of \LaTeX\ Class Files,~Vol.~14, No.~8, August~2015}%
{Shell \MakeLowercase{\textit{et al.}}: Bare Demo of IEEEtran.cls for IEEE Journals}



\maketitle

\begin{abstract}
This paper introduces the Emirates Multi-Task (EMT) dataset, designed to support multi-task benchmarking within a unified framework. It comprises over 30,000 frames from a dash-camera perspective and 570,000 annotated bounding boxes, covering approximately 150 kilometers of driving routes that reflect the distinctive road topology, congestion patterns, and driving behavior of Gulf region traffic. The dataset supports three primary tasks: tracking, trajectory forecasting, and intention prediction. Each benchmark is accompanied by corresponding evaluations: (1) multi-agent tracking experiments addressing multi-class scenarios and occlusion handling; (2) trajectory forecasting evaluation using deep sequential and interaction-aware models; and (3) intention prediction experiments based on observed trajectories. The dataset is publicly available at \href{http://avlab.io/emt-dataset}{avlab.io/emt-dataset}, with pre-processing scripts and evaluation models at \href{http://github.com/AV-Lab/emt-dataset}{github.com/AV-Lab/emt-dataset}.
\end{abstract}


\IEEEpeerreviewmaketitle

\section{Introduction}

\begin{figure}[t!]
\centering
\includegraphics[width=0.49\textwidth]{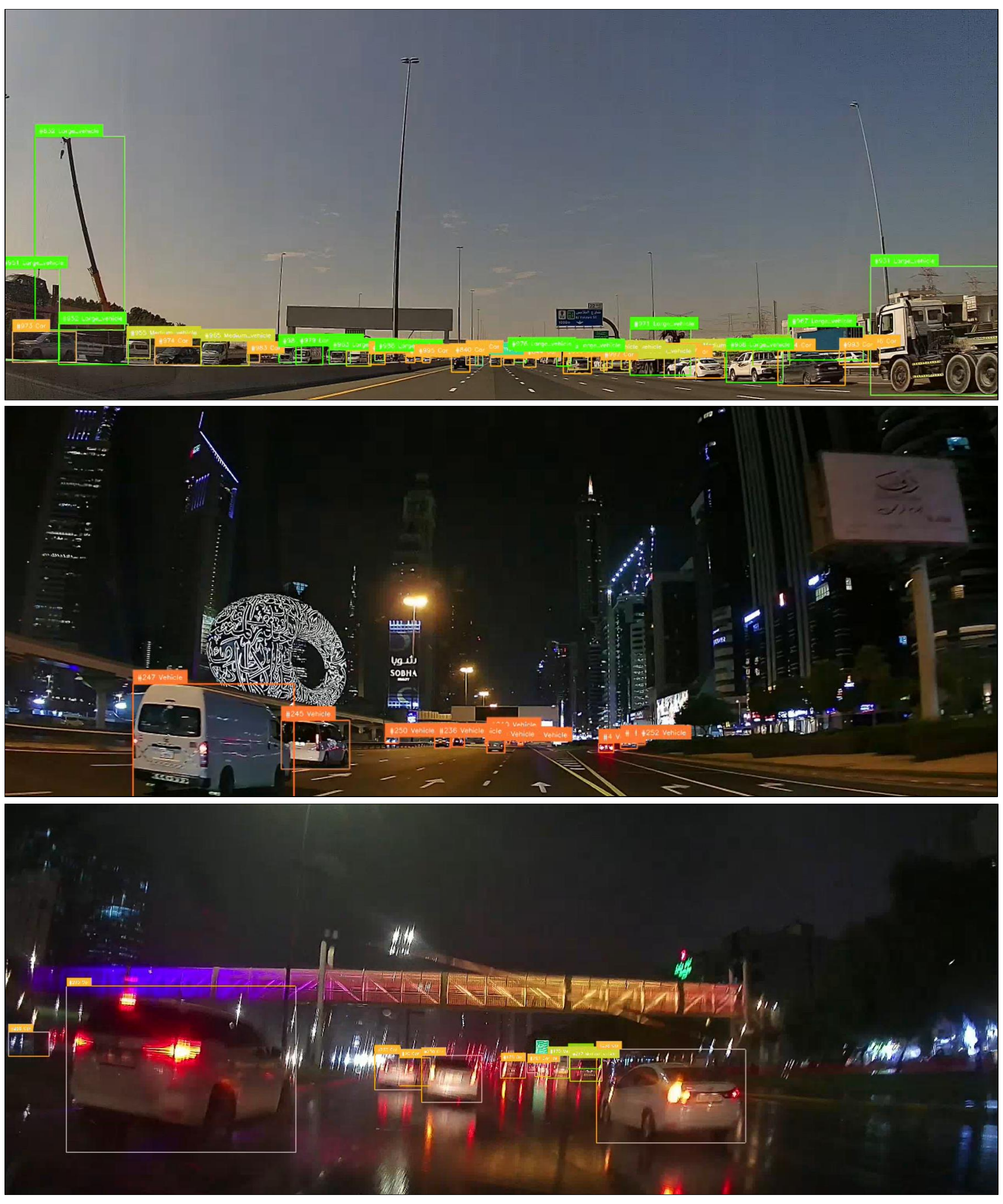}
\caption{Samples from EMT dataset capturing highway scenarios in day and night time, clear and rainy weather.}
\label{fig:sample}
\end{figure}

\begin{table*}[t]
\centering
\caption{Comparison with existing camera-based trajectory–intention datasets highlighting that EMT offers the most extensive set of annotated intentions for both vehicles and pedestrians, along with the most diverse range of traffic scenarios.}
\label{tab:dataset_comparison}
\begin{tabular}{|l|c|c|c|c|c|c|}
\hline
\textbf{Dataset} & \textbf{Region} & \textbf{Tasks} & \textbf{Traffic Participants} & \textbf{Intention Labels} & \textbf{Traffic Scenarios} \\
\hline 
HighD \cite{10.1109/ITSC.2018.8569552} & Germany &  \makecell{-}  & vehicles & \makecell{free-driving, vehicle following, \\ critical maneuver, lane-change} & highway \\ \hline 
JAAD \cite{Kotseruba2016JointAI} & \makecell{USA \\ Canada \\ Eastern Europe} &  \makecell{Intention Prediction \\ Behavior Detection}  & \makecell{pedestrians \\ bicyclists \\ motorcyclists \\ other drivers} & \makecell{crossing, stopped, moving-fast, \\ moving-slow, speed-up, slow-down} & \makecell{intersections \\ urban} \\ \hline
PIE \cite{9008118}  & Canada & \makecell{Trajectory Prediction \\ Intention Prediction \\ Trajectory Prediction \\ conditioned on Intention} & pedestrians & cross, not-cross & urban \\ \hline
STIP \cite{9013045} & USA &  \makecell{Intention Prediction}  & pedestrians & cross, not-cross & urban \\ \hline
EMT (Ours) & \textbf{UAE} & \makecell{Detection \\ Tracking \\ Trajectory Prediction \\ Intention Prediction}  & \makecell{vehicles \\ pedestrians \\ motorbikes \\ cyclists} &  \textbf{\makecell{turn-right, turn-left, merge-right, \\ merge-left, braking, stopped, \\ lane-keeping, reversing, walking, \\ crossing, waiting-to-cross}} & \textbf{\makecell{intersections \\ roundabouts \\ urban \\ highway}} \\
\hline
\end{tabular}
\end{table*}

\IEEEPARstart{A}{s} autonomous driving systems mature, the research community increasingly focuses on the robustness of autonomous vehicle behavior under failures or inaccuracies in different components of the autonomy stack \cite{10.5555/3600270.3602131, 10.1145/3576841.3585930}. Two critical modules influencing both motion planning and overall system safety are perception and prediction, with the latter inherently dependent on the output of the former. Beyond the interdependence between modules, tasks within each module are also tightly coupled. In perception block, detection and tracking are interlinked, i.e., errors in detection propagate to tracking \cite{Geiger2012AreWR, nuscenes2019, 9709630}. In the prediction module, trajectory forecasting and behavioral inference (e.g., pedestrian crossing intention or vehicle maneuvers) are strongly correlated with the tracking perfomance. These interdependencies underscore the need for datasets that support not only task-specific evaluation using ground truth inputs, but also multi-task evaluation where performance dependencies between tasks are preserved. 

In addition to core tasks, recent datasets increasingly incorporate semantic attributes to bridge the gap between raw perception and high-level scene understanding. One such attribute is the intention or maneuver of traffic participants. Human drivers naturally interpret these cues, e.g., a turn signal or lateral drift suggests a lane change for a vehicle. These anticipations, formalized as intentions, shape trajectories and inform collision avoidance. Yet, intention-labeled datasets \cite{10.1109/ITSC.2018.8569552, Kotseruba2016JointAI, 9008118, 9013045, xue2019blvdbuildinglargescale5d, 9710155} remain limited, including joint evaluation of intention and trajectory. Capturing this coupling can facilitate more anticipatory and improve trajectory forecasting accuracy.

Lastly, the ability of data-driven models to generalize across diverse road environments and conditions is essential for safe autonomous operation, yet remains a persistent challenge. Robust generalization requires training on datasets that span a wide range of traffic scenarios. However, the datasets remain scarce, since large-scale annotations are costly and require immense human labor, even with semi-automated annotation pipelines. Additionally, the existing datasets provide coverage of regions such as the USA \cite{nuscenes2019, Houston2020OneTA, Argoverse, Argoverse2, TrustButVerify}, Europe \cite{Geiger2013IJRR, Liao2022PAMI}, and parts of Asia including China and Singapore \cite{10.1609/aaai.v33i01.33016120, nuscenes2019}. The distribution of collected remains geographically uneven, leaving a gap to collect the data from unrepresented regions.

To contribute toward addressing the discussed gaps, we introduce EMT (Emirates Multitask Dataset) --- a dataset constructed from frontal camera footage and designed to support multi-task evaluation. EMT enables consistent benchmarking of object detection, tracking, trajectory forecasting, and intention prediction on temporally aligned data. This structure allows researchers to evaluate each task independently and to examine cross-task dependencies, such as how perception-level errors propagate and affect downstream modules. The dataset is collected in the United Arab Emirates (UAE), a region characterized by a wide range of driving scenarios spanning expansive highways, urban centers, and complex intersections. Specifically, EMT comprises the following task-specific datasets.

The \textbf{tracking benchmark dataset} is designed to evaluate the ability of algorithms to accurately identify and maintain consistent object tracking over time in a complex driving environment for vehicles, pedestrians, cyclists, and motorbikes. The benchmark is designed to test tracking models under varying levels of traffic congestion and frequent lane changes. The dataset contains 8,806 unique tracking IDs, including 8,076 vehicles, 568 pedestrians, 158 motorbikes and 14 cyclists, and with a mean tracking duration of 6.5 seconds.

The \textbf{trajectory prediction benchmark dataset} challenges forecasting models to predict trajectories in heterogeneous traffic and to generalize effectively across scenarios involving interacting agents, forecasts in large intersections and roundabouts, and multi-agent dynamics. The dataset contains 4,821 unique agents, including 4,347 vehicles and 386 pedestrians. Details on the size of the dataset, splits into past trajectories and prediction horizons, as well as the shifting window approach, are provided in Section \ref{sec:dataset}.

The \textbf{intention prediction benchmark dataset} is designed to test models ability to infer the likely actions of surrounding agents based on their past trajectories. For vehicles, the dataset contains maneuver labels: turn left/right, keep lane, merge left/right, brake, stop and reverse and for pedestrians: waiting to cross, crossing, walking (on pavements or sidewalks), and stopping (e.g., waiting at bus stops). These annotations aim to provide a comprehensive understanding of pedestrian and vehicle behaviors. In total the dataset contains 4,921 sequences of agents trajectories with intentions. 

Each task-specific dataset is accompanied by baseline models. For cross-task analysis, we include experiments demonstrating the effect of detection quality on tracking performance. Our key contributions are as follows: 1) a diverse visual dataset that supports multi-task benchmarking, enabling both per-task evaluation and analysis of tasks interdependencies critical for real-world deployment; 2) a benchmark for evaluating the generalization of models trained on data from other regions, offering the resource for assessing performance in underrepresented Gulf driving environments.

The structure of the paper is as follows. Section \ref{sec:rel_work} reviews related work, focusing on existing datasets and models for multi-agent tracking, intention prediction, and trajectory prediction. Section \ref{sec:dataset} outlines the data collection methodology, explains the annotation process, and summarizes the characteristics of the collected dataset. Section \ref{sec:eval_protocol} outlines the evaluation protocol, lists the utilized evaluated models, and presents the results of the conducted experiments for all three datasets. Finally, Section \ref{sec:conclusion} concludes the paper with a summary of findings and directions for future research. 

\section{Related Work}
\label{sec:rel_work}

Related work is summarized through an overview of autonomous driving datasets, as well as models for multi-object tracking, trajectory forecasting, and intention prediction.

\subsection{Autonomous Driving Datasets}
Evaluating the effectiveness of autonomous driving algorithms requires testing across a diverse range of scenarios to assess performance under various conditions. Each scenario presents unique challenges and variability that models must navigate to be considered viable for real-world application. In recent years, numerous datasets, such as KITTI \cite{Geiger2013IJRR}, KITTI-360 \cite{Liao2022PAMI}, ApolloScape \cite{10.1609/aaai.v33i01.33016120}, PIE \cite{9008118}, NuScenes \cite{nuscenes2019}, WAYMO \cite{9709630}, Argoverse I \cite{Argoverse}, Argoverse II \cite{Argoverse2,TrustButVerify}, and Lyft Level 5 \cite{Houston2020OneTA}, have been introduced to support research in tasks such as detection, segmentation, tracking, prediction, intention prediction, and planning.

In the context of multi-task learning, BDD100K \citep{9156329} is a large-scale dataset comprising approximately 100K driving videos, designed to support multi-task training across 10 vision tasks relevant to autonomous driving, including object detection, tracking, semantic segmentation, lane marking, and drivable area detection. Among trajectory prediction datasets, nuScenes \cite{nuscenes2019}, Argoverse \cite{Argoverse,Argoverse2}, and WAYMO \cite{9709630} motion forecasting datasets are commonly used for state-of-the-art model evaluation. Focusing on interactions between traffic participants, the INTERACTION dataset \cite{zhan2019interactiondatasetinternationaladversarial} captures complex environments such as roundabouts, intersections, and highways, offering scenarios critical for interaction-aware trajectory prediction.

Related to intention prediction datasets, Table~\ref{tab:dataset_comparison} includes a comparison between the proposed EMT and existing camera-based intention datasets \cite{10.1109/ITSC.2018.8569552, Kotseruba2016JointAI, 9008118, 9013045}. In addition, among multi-sensor datasets, the BLVD dataset \cite{xue2019blvdbuildinglargescale5d} was collected in China and designed primarily for tasks such as interactive event recognition and intention prediction. Events in BLVD combine landmark estimation with fine-grained intention prediction. The dataset covers vehicles, pedestrians, and riders (cyclists and motorcyclists), and provides detailed intention labels: 13 types for vehicles, 8 for pedestrians, and 7 for riders, enabled by the sensor fusion setup that allows for accurate velocity estimation. Similarly, LOKI dataset \cite{9710155} was collected in Japan using a multi-sensor rig. The dataset targets long-term and short-term intention prediction in dense urban scenes. It provides both frame-wise labels (e.g., stop, lane change, cut-in, walk, wait) and long-term goal annotations vehicles and pedestrians.

The designed EMT dataset complements existing camera-based intention datasets by offering a more fine-grained set of intention labels and greater diversity in traffic scenarios. Among multi-sensor intention datasets, EMT also fills a geographic gap by contributing data collected from a previously underrepresented region. By capturing diverse driving conditions typical of Arab Gulf-region traffic, EMT introduces a unique context that supports the development of models with improved predictive accuracy under culturally and regionally distinct driving behaviors.

\subsection{Multi-object Tracking}
Multi-object tracking (MOT) methods encompass a variety of approaches, including tracking-by-detection \cite{8296962, 9010033} and joint detection and tracking \cite{zhang2021fairmot, 10.1007/978-3-030-58621-8_7}. Tracking-by-detection methods first detect objects in each frame and then associate them across frames. SORT \cite{8296962} serves as a lightweight and efficient baseline in this category, combining object detections with a Kalman Filter for motion prediction and the Hungarian algorithm for data association. Tracktor \cite{9010033}, on the other hand, eliminates the need for explicit data association by reusing regression heads in object detectors to refine and propagate tracks. Joint detection and tracking approaches integrate object detection and tracking into a unified pipeline. FairMOT \cite{zhang2021fairmot} optimizes a shared CNN backbone for simultaneous detection and re-identification, operating in crowded scenes. The Joint Detection and Embedding (JDE) tracker \cite{10.1007/978-3-030-58621-8_7} generates detections and embeddings in real-time by sharing features between tasks and optimizing with a multi-task loss.

Kalman filter-based trackers \cite{BoT-SORT, bytetrack, 8296962, 10406854,10160328} use motion prediction to refine trajectory estimation and handle missing detections. OC-SORT \cite{cao2023observation} builds upon SORT by introducing motion consistency constraints, addressing fragmentation in crowded scenarios. ByteTrack \cite{bytetrack} employs Kalman Filters to ensure accurate motion estimation during association.

\subsection{Trajectory Prediction Methods}

The field of trajectory prediction is extensively studied and encompasses various methodologies, including physics-based models, probabilistic approaches, and learning-based techniques. The presented related work focuses on deep-learning methods relevant to our evaluation framework: LSTM-based models, Transformer-based architectures, and Graph Neural Networks (GNNs).

\subsubsection{LSTM-based methods}

LSTM-based models have proven effective in capturing the temporal dependencies inherent in trajectory data. Early approaches, such as Social-LSTM \cite{7780479}, introduced the concept of social pooling to model interactions among agents, establishing a foundation for socially-aware trajectory prediction. Subsequent variants, including SSCN-LSTM \cite{Varshneya2017HumanTP} and SR-LSTM \cite{zhang2019srlstmstaterefinementlstm}, extended these capabilities by integrating spatial context and message-passing mechanisms, thereby enhancing their ability to represent dynamic interactions in multi-agent environments.

Several works have improved LSTM architectures by integrating environmental and spatial factors. Scene-LSTM \cite{Manh2018SceneLSTMAM} utilized a grid-based scene representation to capture interactions between agents and static objects, while MX-LSTM \cite{Hasan2018MXLSTMMT} combined trajectory data with head pose estimations to model multimodal behavior. GC-VRNN \cite{Xu2023UncoveringTM} extended this concept further by incorporating graph-based representations to manage incomplete data and complex temporal dependencies. TraPHic \cite{8954462} demonstrated the effectiveness of combining CNNs with LSTMs to handle heterogeneous traffic conditions.

Architectural advancements have also enabled multimodal and hierarchical designs. For example, stacked LSTMs \cite{10.1109/IVS.2018.8500493}, hierarchically process trajectory data to predict maneuver-specific behaviors, while encoder-decoder frameworks \cite{Zyner2018NaturalisticDI}, utilize Gaussian Mixture Models for probabilistic trajectory predictions. Other works, like STS-LSTM \cite{zhang_lstm_2023} and Highway-LSTM \cite{8317913} highlight the adaptability of LSTMs in specialized use cases, such as spectral trajectory modeling or highway motion forecasting.

\subsubsection{Transformers} Transformer-based models are highly effective in capturing long-range dependencies, which are crucial for understanding agent motion over time. STAR \cite{yu2020spatiotemporalgraphtransformernetworks} leverages Transformers to model individual pedestrian dynamics and graph-based spatial Transformers for crowd interactions, using alternating layers for joint spatio-temporal modeling. Similarly, GA-STT \cite{Zhou2022GASTTHT} employs cross-attention to fuse spatial and temporal embeddings, effectively capturing both individual and group-level motion features. HiVT \cite{9878832} adopts a two-stage approach, with the first stage focusing on extracting local context and the second stage integrating global interactions.

For social interaction modeling, AgentFormer \cite{9710708} utilizes an agent-aware Transformer with tailored attention mechanisms for intra-agent and inter-agent interactions. LatentFormer \cite{Amirloo2022LatentFormerMT} introduces hierarchical attention to capture social interactions and employs Vision Transformers to extract contextual scene features. Several models integrate GNNs with Transformers, such as Graph-Based Transformers \cite{Li2022GraphbasedST, zhang_2022, 10504962}, which model structured representations of agent-agent and agent-environment relationships, highlighting the synergy between graph representations and Transformer architectures for structured interaction data.

In the context of multimodal integration, the literature explores combining heterogeneous input modalities, such as map information, bounding boxes, and scene semantics. mmTransformer \cite{9577819} integrates a motion extractor for past trajectories, a map aggregator for road topology, and a social constructor to model agent interactions. MacFormer \cite{Feng_2023} builds upon this by incorporating map constraints and Crossmodal Transformers \cite{9812226} further advance this approach by fusing cross-relation features between modality pairs, supported by a modality attention module.

\subsubsection{GNN-based methods}
GNN models represent traffic scenes as graphs, where nodes correspond to agents and edges encode relationships, typically based on proximity. Graph Convolution Networks (GCN)-based models focus on spatial relationships and dynamic interaction modeling. GRIP \cite{8917228} alternates between temporal convolutional layers and graph operations to encode motion features and spatial interactions, while its extension, GRIP++ \cite{Li2019GRIPEG}, enhances this by incorporating dynamic edge weights and scene context. LaneGCN \cite{liang2020learning} integrates HD map data using lane graphs, and Trajectron++ \cite{10.1007/978-3-030-58523-5_40} combines GCNs with recurrent architectures to process spatiotemporal graphs enriched with semantic features.

Another line of work relies on Graph Attention Networks (GAT). GAT-based models use attention mechanisms to prioritize influential interactions dynamically. Social-BiGAT \cite{e294141389194a54a05536938fcdd509} combines GATs with generative frameworks for multimodal forecasting. GATraj \cite{CHENG2023163} introduces a Laplacian mixture decoder to enhance prediction diversity while modeling spatial-temporal dependencies, and GraphTCN \cite{Wang2020GraphTCNSI} integrates GATs with temporal convolutional networks (TCNs) to efficiently capture long-term dependencies.

Hierarchical and hybrid approaches combine multiple techniques for comprehensive modeling. MFTraj \cite{Liao2024MFTrajMB} uses dynamic geometric graphs and adaptive GCNs to model spatiotemporal dependencies. GOHOME \cite{10.1109/ICRA46639.2022.9812253} and MTP-GO \cite{Westny2023MTPGOGP} adopt graph-based methods for agent-map interactions, with MTP-GO employing neural ODEs to manage dynamic motion constraints. EqMotion \cite{10205349} ensures interaction invariance and geometric equivariance, providing stable and consistent trajectory predictions.

\subsection{Intention Prediction Methods}

Intention-aware models represent an important direction in trajectory prediction, aiming to enhance accuracy by incorporating agents’ maneuver intentions. These methods \cite{7487409, Deo2018HowWS} treat maneuvers as short-term, goal-driven decisions that consist of a sequence of continuous states working toward a global objective. Maneuvers are typically categorized into lateral decisions, such as lane-keeping, lane-changing, or turning, and longitudinal decisions, such as maintaining speed, accelerating, or braking. By integrating maneuver awareness, these models introduce an intermediate reasoning layer that informs predictions with planning-based logic.

Several LSTM-based models have demonstrated success in intention prediction. Occupancy-LSTM \cite{Kim2017ProbabilisticVT} generates occupancy grid maps by modeling surrounding vehicle motions, capturing likely maneuvers such as lane changes. Similarly, Zyner et al. \cite{10.1109/IVS.2017.7995919} and Phillips et al. \cite{Phillips2017GeneralizableIP} employ LSTMs to predict driver intentions, using features like heading, position, and velocity, with a particular focus on intersections. MX-LSTM \cite{Hasan2018MXLSTMMT} extends this capability by incorporating head pose data, providing an additional layer of contextual awareness to better infer maneuver intentions.

\section{Dataset}
\label{sec:dataset}

In this section, the key aspects of the EMT dataset are introduced, including the data collection process, design methodology, and annotations format. 

\begin{table*}[t]
\centering
\caption{Object classes description and statistics.}
\begin{tabular}{|p{3cm}|p{10cm}|c|c|}
\hline
\textbf{Class} & \textbf{Description} & \textbf{\makecell{Number of \\ Bounding Boxes}} & \textbf{\makecell{Number of \\ Agents}} \\  \hline
Pedestrian & An individual walking on foot.  & 24,574 & 568 \\ \hline
Cyclist & Any bicycle or electric bike rider. & 594 & 14  \\ \hline
Motorbike & Includes motorcycles, bikes, and scooters with two or three wheels. & 11,294 & 159 \\ \hline
Car & Any standard automobile. & 429,705 & 6,559 \\ \hline
Small motorized vehicle & Motorized transport smaller than a car, such as mobility scooters and quad bikes. & 767 & 13 \\ \hline
Medium vehicle &  Includes vehicles larger than a standard car, such as vans or tractors. & 51,257 & 741 \\ \hline
Large vehicle & Refers to vehicles larger than vans, such as lorries, typically with six or more wheels. & 37,757 & 579 \\ \hline
Bus & Covers all types of buses, including school buses, single-deck, double-deck. & 19,244 & 200 \\ \hline
Emergency vehicle & Emergency response units like ambulances, police cars, and fire trucks, distinguished by red and blue flashing lights. & 1,182 & 9 \\ \hline
\multicolumn{2}{|r|}{\textit{\textbf{Overall:}}} & \textbf{576,374} & \textbf{8,842} \\ \hline
\end{tabular}
\label{tab:classes_desc}
\end{table*}

\begin{figure*}[h!]
\centering
\includegraphics[width=1.0\textwidth]{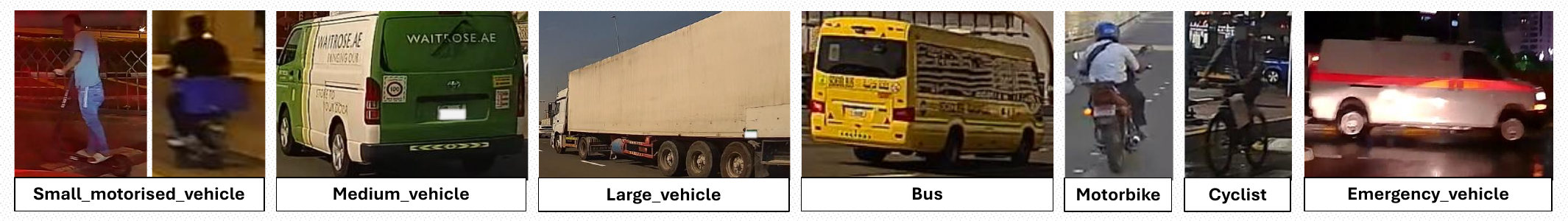}
\caption{Samples of annotated agents, including small motorized vehicles, medium and large vehicles, emergency vehicles, buses, motorbikes (comprising motorbike and rider), and cyclists (comprising bicycle and rider).}
\label{fig:agents}
\end{figure*}

\subsection{Data collection}
The EMT dataset was collected in two major cities in the UAE - Abu Dhabi and Dubai, as well as on the roads connecting these cities. Data collection was conducted using two vehicles equipped with front-facing cameras. Each vehicle was outfitted with a VANTRUE 3-Channel Dash Camera, recording at 1080P Full HD. The video footage captures a variety of road topologies common in the region, including highways, roundabouts, bridges, city junctions, intercity highways, and narrow urban streets. The dash cameras recorded video sequences at a frame rate of 30 fps, saved as 3-minute clips. For the annotation process, frames were extracted at 10 fps. No pre-processing was applied to remove flashes or blurring caused by bumpy sections along some routes, as this aspect was left for future research to integrate as part of model design. Videos were carefully selected to represent a range of weather conditions, times of day, and scene types. This selection includes recordings during bright daylight, evening, and nighttime, under clear and rainy weather conditions. This diverse collection ensures the inclusion of the most common scenes encountered in the Arab Gulf region. Each video clip lasts between 2.5 and 3 minutes, totaling approximately 57 minutes of video data in the dataset.

\subsection{Data Annotations}
Every significant object and actor within each frame is annotated, including vehicles, pedestrians, small motorized vehicles, motorbikes, and cyclists. For objects classification and description we relied on convention presented in \cite{9712346}. Small motorized vehicles refer to any motorized vehicle smaller than a car, such as mobility scooters or quad bikes. Medium vehicles include those larger than a car, such as vans, while large vehicles refer to lorries, typically characterized by having six or more wheels. Emergency vehicles include ambulances, police cars, and fire engines equipped with red and blue flashing lights, excluding vehicles with yellow flashing lights, such as road maintenance vehicles. Each object is assigned an intention label based on its observed behavior. Table \ref{tab:classes_desc} provides for each class description, number of bounding boxes and number of agents throughout the whole dataset.

For vehicles, the intention reflects the current maneuver being performed, while for pedestrians, it represents their activity. The most common maneuver for vehicles is lane-keeping, where the vehicle maintains its current lane and follows a steady trajectory without deviation. The labels "merge left" and "merge right" are used when a vehicle moves to an adjacent lane, merges, or exits a main road. Turning maneuvers, such as "turn left" and "turn right," are assigned when a vehicle turns at an intersection or road junction, typically involving a reduction in speed and a change in direction. Braking indicates deceleration, often observable through activated brake lights in image-based data. Reversing refers to backward movement, which is commonly observed in parking areas or near road edges. Stop is used when a vehicle has come to a complete halt, such as at an intersection waiting for a green light or in parking areas. For pedestrians, the intention label walking denotes movement along a road or sidewalk at a steady pace without any apparent intent to cross the road. Waiting to cross is used when a pedestrian is facing oncoming traffic, signaling potential preparation to cross. Crossing applies when a pedestrian moves across the road, whether at junctions, designated crossings, or random locations, as long as the movement involves passing in front of a vehicle. The stop label indicates that the pedestrian is stationary, such as standing on a pavement or waiting at a bus stop. This annotation approach provides a comprehensive foundation for analyzing and understanding dynamic interactions between agents within road scenes.

\begin{figure*}[h!]
\centering
\includegraphics[width=1.0\textwidth]{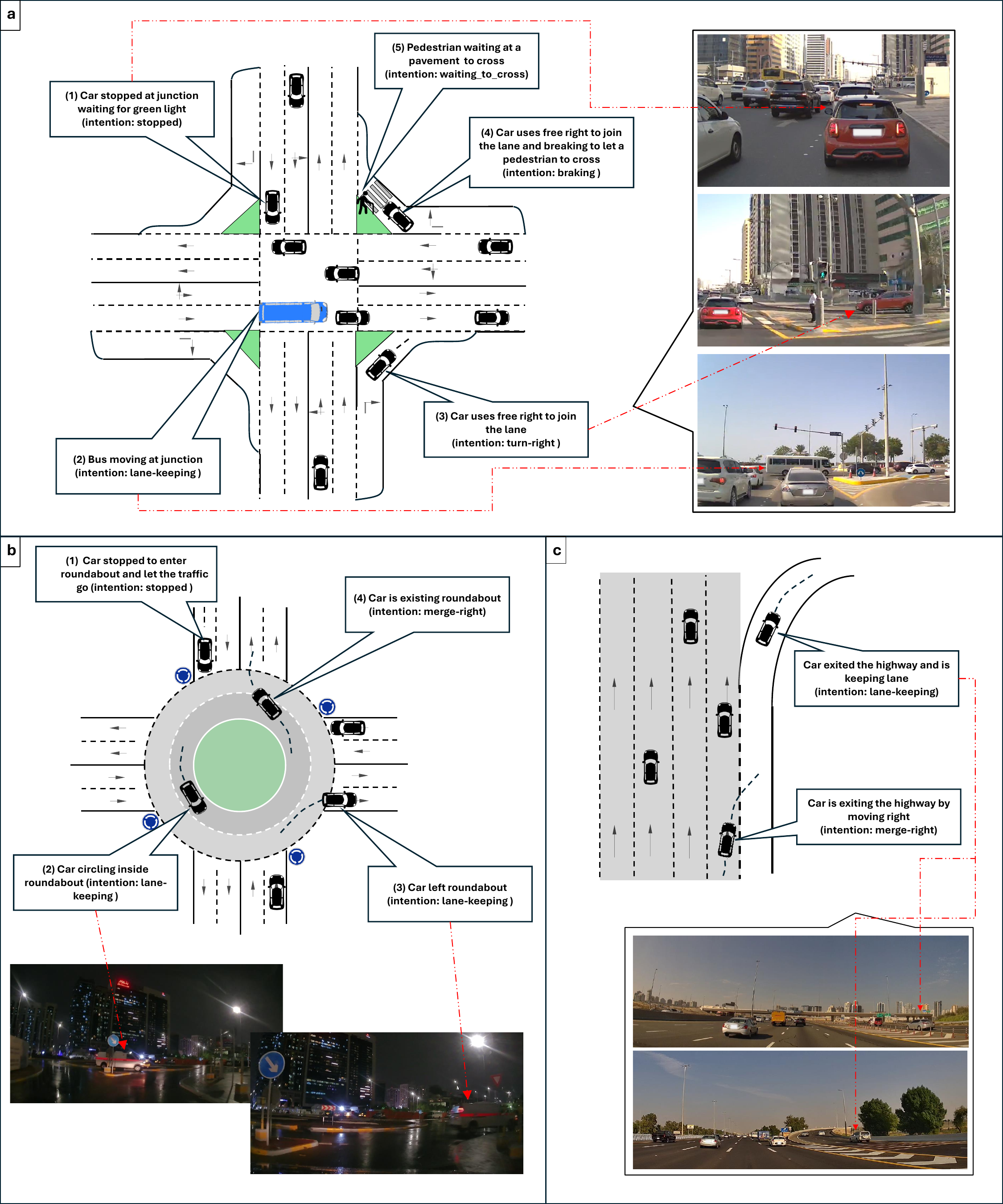}
\caption{Common traffic scenarios in the UAE include: (a) large city junctions, often featuring unique elements like free right turns to reduce congestion, (b) roundabouts with vehicles navigating through various stages such as approaching, circulating, and exiting, and (c) highways with exits, where vehicles transition smoothly by merging into designated lanes and exiting the main road.}
\label{fig:junction}
\end{figure*}

\begin{table*}[t]
\centering
\caption{Description of Collected Videos Train and Test Splits}
\begin{tabular}{|c|p{5.5cm}|p{1.5cm}|p{1.5cm}|p{1.5cm}|p{5.5cm}|}
\hline
& \textbf{Videos} & \textbf{Number of Bounding Boxes} & \textbf{Number of Frames} & \textbf{Resolutions}  & \textbf{Lightning and Weather Conditions, \:\:\:\: \:\:\:\: Road Elements} \\ \hline
\multirow{1}{*}[1em]{\rotatebox[origin=c]{90}{\textbf{Train Set}}} & \makecell{video\_115533, video\_115833, video\_141733, \\ video\_161205, video\_151901, video\_143739, \\ video\_142911, video\_054604, video\_054907, \\  video\_131333, video\_204647, video\_214547, \\  video\_210906, video\_220047, video\_174340 } & \centering  471,649 & \centering 26,360 & \makecell{2560x1440 \\ 896x672 \\ 3840x2160}   & \makecell[l]{day, evening \\ clear weather, heavy rain \\ driving in city, highways, junctions, at parking, \\ at roundabouts,  merging main road, bridges, \\ between cities}   \\ \hline

\multirow{1}{*}[1em]{\rotatebox[origin=c]{90}{\textbf{Test Set}}} & \makecell{video\_122233, video\_204347, video\_125233, \\ video\_155425, video\_160325  } & \centering  98,049 & \centering 8,530 & \makecell{2560x1440 \\ 3840x2160}   & \makecell[l]{day, evening \\ clear weather, rainy \\ driving in city, highways, at roundabouts,  \\ merging main road, bridges, between cities}   \\ \hline

\hline
\end{tabular}
\label{tab:dataset}
\end{table*}

\begin{table*}[t]
\centering
\caption{Trajectory Prediction Settings: The notation 30/60 indicates the duration of the past trajectory and prediction horizon, measured in frames. This setup corresponds to 3 seconds of observed data followed by 6 seconds of predicted trajectory. Each row corresponds to different sliding window setting and shows the size of train and test datasets.}
\begin{tabular}{|p{3cm}|c|c|c|c|c|c|}
\hline
 & \textbf{10/10, 1s/1s} & \textbf{10/20, 1s/2s} & \textbf{20/10, 2s/1s} & \textbf{20/20, 2s/2s} & \textbf{20/30, 2s/3s}  & \textbf{20/60, 2s/6s} \\ \hline
sliding window = 1 & 268,504/71,097 & 234,969/64,007 & 234,969/64,007 & 210,709/58,637 & 191,772/54,335 & 151,218/44,989 \\ \hline
sliding window = 3 & 90,897/71,097 & 79,261/64,007 & 79,261/64,007 & 70,931/58,637 & 64,508/54,335 & 50,779/44,989  \\ \hline
sliding window = 5 & 55,408/71,097 & 48,159/64,007  & 48,159/64,007 & 43,020/58,637 & 39,057/54,335& 30,693/44,989  \\ \hline
\end{tabular}
\label{tab:prediction_datasets}
\end{table*}

The most common road topologies for the region are illustrated in Figure \ref{fig:junction}. Figure \ref{fig:junction}(a) depicts a typical large city junction common in the region. A key feature of such junctions, compared to those in other areas, is the inclusion of a free right turn to alleviate traffic congestion. Cases (3) and (4) highlight this scenario: vehicles can make a free right turn from the traffic lights if no pedestrians are present and the turn does not interfere with other traffic. Otherwise, they must wait to complete the turn and merge safely onto the intended road. For clarity, the figure also includes other typical scenarios: (1) a car waiting at the junction for the green light; (2) a bus crossing the junction from left to right while staying in the same lane; (3) a car performing a free right turn to merge into a lane; (4) a car slowing down to allow a pedestrian to cross, tagged with the action "braking"; and (5) a pedestrian standing on the pavement, waiting for the car to stop before crossing safely. Figure \ref{fig:junction}(b) schematically depicts a roundabout with eight access points, showing vehicles in various stages: approaching, circulating inside, and exiting. When the ego vehicle intends to enter the roundabout, it must monitor agents in the outer or both lanes, depending on its intended exit, to avoid collisions. Each vehicle within the roundabout is labeled with an intention, such as "stopped" if it is waiting to enter, "keep-lane" in case of moving within the roundabout, and "merge-right" when exiting the roundabout. Lastly, Figure \ref{fig:junction}(c) represents a typical highway layout for the region. It illustrates cases where a vehicle exits the highway and gradually separates from the main road. When a vehicle begins exiting by curving right and changing lanes, its intention is labeled as "merge-right." Once the vehicle has exited, the label remains "keep-lane" until the vehicle disappears from the ego vehicle’s field of vision. All scenarios are supplemented with corresponding frames from the dataset.

\subsection{Annotations Format}

For Multi-Object Tracking, annotations are provided in two formats: GMOT and KITTI. Each video’s annotation file includes objects detected across frames, with bounding boxes defined by the $x$ and $y$ coordinates of the top-left and bottom-right corners. These files also include the agent's class and consistent tracking IDs for all objects. For trajectory and intention prediction datasets, the annotation format closely follows the PIE dataset \cite{9008118}. Each video’s annotation file contains all recorded objects, along with their unique IDs, agent classes, sequences of frames in which they appear, and corresponding sequences of bounding boxes. In intention prediction, each object additionally has an ``intention" attribute that stores the sequence of intentions throughout its entire trajectory:

\begin{lstlisting}
[
    {
        "id": 4,
        "class": "Car",
        "frames": [1, 2, 3, 4],
        "bbox": [
            [1308.9649, 1031.4597, 1349.817, 1057.7218],
            [1305.2132, 1030.2091, 1346.0653, 1056.4712],
            [1303.1289, 1029.3754, 1340.6461, 1056.4712],
            [1304.3794, 1031.4597, 1338.9787, 1057.3049]
        ],
        "intention": ["lane-keeping", "lane-keeping", "lane-keeping", "lane-keeping"]
    }, ...
]
\end{lstlisting}

To enhance usability, we supplement the annotations with a parsing script capable of generating custom settings by varying the size of past trajectories, prediction horizon, and overlaps. The script iterates through all objects and segments their trajectories into samples of past and future trajectories using a sliding window approach. For example, if the sliding window is set to 3 frames, the first sample will be generated starting from \textit{frame\_1}, and the next from \textit{frame\_4}. The default format is based on past observations and is structured as: \([past\_trajectory]\ [future\_trajectory]\). Table \ref{tab:prediction_datasets} presents various settings along with the sizes of the training and test datasets. Additionally, we provide a scene-centered generation option, with further details available in Section \ref{sec:Prediction}.

While all tasks are derived from the same video sequences and temporally aligned scenes, task-specific filtering is applied during preprocessing. For instance, parked vehicles are retained in the tracking benchmark to ensure that models learn to distinguish static objects and maintain consistent identities. In contrast, such objects are excluded from trajectory forecasting, which focuses on modeling agent dynamics. These differences necessitate task-specific object enumeration, resulting in non-shared object IDs across tasks. Nonetheless, the datasets share the same image streams and maintain consistent training and testing splits, ensuring contextual coherence for cross-task dependencies.

\subsection{EMT Statistics and Annotations}

The dataset contains 20 videos, encompassing both day and night conditions, as well as a variety of weather scenarios, including rainy and clear weather. Table \ref{tab:dataset} provides details for each video, including the city, time of driving, weather conditions, and road topologies. In total, the dataset contains approximately 570,000 bounding boxes. The Table \ref{tab:dataset} also reports the mean density of agents per frame for each video, ranging from a minimum of around 6 agents per frame to a maximum of 30 agents, highlighting the density and diversity of the dataset.

The annotations were created using the Basic.ai tool --- annotation platform equipped with tracking algorithms, propagating bounding boxes across frames while maintaining consistent unique identifiers. Its robust tracking functionality was a key reason for its selection, significantly streamlining the annotation process. On average, each video required approximately 70 hours of annotation, completed by professional human annotators.
\section{Evaluation}
\label{sec:eval_protocol}

This section presents the experimental results conducted on all three task-specific datasets. For each set of experiments, we provide a description of the evaluated models, specify the evaluation protocol, outline the implementation details, and report the results. 

\begin{table*}[t!]
\centering
\caption{Off-shelf vs Fine tuned YOLOX-L detection results (the best results are highlighted in bold)}
\begin{tabular}{|p{2cm}|c|c|c|c|c|c|c|c|c|c|}
\hline
\multirow{2}{*}{\textbf{Object Class}} & \multicolumn{5}{c|}{\textbf{YoloX Off-shelf}} & \multicolumn{5}{c|}{\textbf{YoloX fine-tuned}} \\
\cline{2-11}
& \textbf{Precision↑} & \textbf{Recall↑} & \textbf{F1Score ↑} & \textbf{FP↓} & \textbf{FN↓} & \textbf{Precision↑} & \textbf{Recall↑} & \textbf{F1Score ↑} & \textbf{FP↓} & \textbf{FN↓} \\
\hline
Vehicle & 90.14 & 23.59 & 37.39 & 2,669 & 79,002 & 85.53 & 66.87 & 75.06 & 11,692 & 34,257   \\
Pedestrian & 65.59 & 10.33 & 17.84 & 255  & 4,221 & 82.76 & 73.13 & 77.64 & 717 & 1,265 \\
Motorbike &  96.15 &  4.61 &  8.79 & 4 & 2,071  & 93.12 & 56.70 & 70.48 & 91 & 940  \\
Cyclist & 50.00 & 0.65 & 1.28 & 1 & 153 & 93.63 &  95.45 & 94.53 & 10 & 7 \\
\cline{1-11}
\textbf{overall} & \textbf{89.50} & 22.62 & 36.11 & \textbf{2,929} & 85,447 & 85.53 &\textbf{66.97} & \textbf{75.12} & 12,510 & \textbf{36,469}  \\ 
\hline
\end{tabular}
\label{tab:detection-eval}
\end{table*}

\begin{table*}[t!]
\centering
\caption{Tracking Evaluation Results for six Detector-Tracker Settings (the best results for fine-tuned setting between two trackers are highlighted in bold)}
\begin{tabular}{|p{0.4cm}|p{1.05cm}|c|c|c|c|c|c|c|c|c|c|c|c|}
\hline
\multirow{2}{*}{\textbf{Det.}} & \multirow{2}{*}{\textbf{\makecell{Object \\ Class}}} & \multicolumn{6}{c|}{\textbf{BoT-SORT}} & \multicolumn{6}{c|}{\textbf{ByteTrack}} \\
\cline{3-14}
& & \textbf{MOTA↑} & \textbf{HOTA↑} & \textbf{IDF1↑} & \textbf{FP↓} & \textbf{FN↓} & \textbf{IDs↓} & \textbf{MOTA↑} & \textbf{HOTA↑} & \textbf{IDF1↑} & \textbf{FP↓} & \textbf{FN↓} & \textbf{IDs↓} \\
\hline

\multirow{4}{0.2cm}{\rotatebox{90}{\textbf{\makecell{Ground \\ Truth}}}} 
& Vehicle & 92.0 & 73.9 & 74.8 & 117,353 & 147,168 & 12,992 &  85.4 & 66.9 & 72.5 & 124,039 & 162,065 & 14,044  \\
& Pedestrian & 84.9 & 70.9 & 76.8 & 4,158  & 6,674& 1,156 & 70.6 & 59.8 & 68.0 & 4,676 & 9,493 & 1,742 \\
& Motorbike &  86.8 &  77.1 &  79.6 & 1,542 & 2,800 & 200 & 83.0 & 71.4 & 76.8 & 1,723 & 3,180 & 208  \\
& Cyclist & 93.9 & 77.02 & 89.0  & 50 & 79 &  7 & 73.9 &  60.3 & 75.8 & 61 & 194 &  16\\
\cline{1-14}

\multirow{4}{0.2cm}{\rotatebox{90}{\textbf{\makecell{YoloX \\ Off-shelf}}}} & Vehicle & 20.2 & 25.9 & 29.4 & 20,924 & 443,214 & 1,077 & 19.8 & 25.4 & 29.6 &  25,507 & 441,511 & 1,757 \\
& Pedestrian & 6.6 & 14.1 & 14.0 & 757  & 22,656 &  53 & 6.9 & 13.7 & 13.8 & 721 & 22,689 & 67 \\
& Motorbike & 3.7 & 8.3 & 5.9 & 135 & 10,944 &  10 & 3.7 & 8.5 & 5.9 & 157 & 10,940 & 13  \\
& Cyclist & 0 & 6.5 & 6.4 & 40 & 573 & 1 & 0 & 3.6 & 3.5 & 27 & 583 & 0 \\
\cline{1-14}

\multirow{4}{0.2cm}{\rotatebox{90}{\textbf{\makecell{YoloX \\ fine-tunned}}}} & Vehicle & \textbf{72.7} & \textbf{57.6} &  \textbf{65.8} & \textbf{130,451} & \textbf{211,305} &  \textbf{11,009}  & 67.4 & 55.0 & 64.7 & 132,892 & 217,837 & 12,055 \\
& Pedestrian & \textbf{58.7} & \textbf{50.4} & \textbf{63.0} & \textbf{4,909} & \textbf{11,011} & \textbf{960} & 48.5 & 44.8 & 55.4 & 5,175 & 13,171 & 1,361 \\
& Motorbike & \textbf{63.0} & \textbf{58.1} &  \textbf{68.2} & \textbf{1,255} & \textbf{4,804} & \textbf{127} & 60.7 & 56.2 & 65.9 & 1,377 & 5,063 & 113 \\
& Cyclist & \textbf{64.8} & \textbf{54.5} & \textbf{69.9} & 74 & \textbf{235} & \textbf{6} & 45.8 & 40.0 & 56.3 & \textbf{57} & 339 & 14 \\
\cline{1-14}
\hline 
\end{tabular}
\label{tab:tracking-eval}
\end{table*}

\subsection{Multi-Agent Tacking}
Multi-agent tracking experiments are conducted using Kalman filter-based trackers in a tracking-by-detection setup. Two SOTA trackers are evaluated to assess their ability to handle occlusions and detection errors, particularly when dealing with far objects.

\subsubsection{Trackers}
\begin{itemize}
     \item \textbf{ByteTrack} \cite{bytetrack} is a lightweight real-time multi-object tracking model employing a two-level data association method. ByteTrack processes both high-confidence and low-confidence detections, enabling more robust tracking while maintaining computational efficiency. The method first associates tracklets with high-confidence detections, then matches unmatched tracklets with low-confidence detections to recover lost tracks. 

    \item \textbf{BoT-SORT} \cite{BoT-SORT} is a multi-object tracker that enhances motion-based tracking through camera-motion compensation and an improved Kalman filter state representation. Like ByteTrack, it employs a two-level association strategy. The tracker employs a cascade matching approach that prioritizes motion information for data association.
\end{itemize}

\subsubsection{Evaluation Protocol}
We evaluate the trackers using three different detector configurations: (1) ground-truth detections assuming the presence of perfect detector, providing an upper bound on performance; (2) off-the-shelf YOLO detector, representing readily available solutions; (3) fine-tuned YOLO detector, tuned for our use case. Each detector configuration serves a distinct analytical purpose. Ground-truth detections allow us to evaluate the tracker's association algorithm under ideal conditions, isolating its core tracking capabilities from detection errors. The off-the-shelf YOLO detector helps assess tracker robustness under challenging conditions with missing and erroneous detections. Finally, the fine-tuned YOLO detector represents a more realistic production scenario, where the detector is optimized for the specific domain but still maintains some inherent errors.
We conduct the evaluation using F1-score, Identity Switches and Higher Order Tracking Accuracy (HOTA):
\begin{itemize}
\item \textit{Multi-Object Tracking Accuracy (MOTA)}: Evaluates overall tracking performance by accounting for false positives, false negatives, and identity switches.
\item \textit{False Positives (FP)}: Instances where the tracker incorrectly identifies an object that is not present in the ground truth (ghost detections).
\item \textit{False Negatives (FN)}: Cases where the tracker fails to identify an object present in the ground truth due to detection or tracking errors (missed detections).
\item \textit{Identity Switches (IDs)}: The total number of instances where a tracker incorrectly reassigns identities between objects.
\item \textit{Identity F1 Score (IDF1)}: Measures the tracker's ability to maintain consistent object identities, considering identity switches and fragmentation.
\item \textit{Higher Order Tracking Accuracy (HOTA)}: A comprehensive metric that balances detection and association accuracy, decomposed into detection accuracy (DetA) and association accuracy (AssA) components.

\end{itemize}
To match YOLO's classes, we group EMT object classes into four superclasses: pedestrian, motorbike, cyclist, and vehicle. The vehicle superclass encompasses all other classes in the dataset, including cars, buses, small motorized vehicles, medium and large vehicles, and emergency vehicles. Tracking performance metrics are broke down for each superclass to provide a detailed analysis of tracker behavior across different object categories. The evaluation was done on the full dataset, i.e., including train and test splits.

\subsubsection{Implementation Details}
As an off-the-shelf detector, the YOLOX-L model from ByteTrack was utilized. For the fine-tuned setting, YOLOX-L was trained on a train-test split with an input image resolution of 1280×1280. Since the original implementations of both trackers were designed for single-class pedestrian tracking, we extended their capabilities to support multi-class MOT. Our implementation maintains separate tracking instances for each class, ensuring that object association occurs only between detections of the same class. This preserves the core tracking logic while enabling simultaneous tracking of multiple object categories.

All experiments followed our evaluation protocol using standard tracking metrics. To ensure reproducibility, we maintained consistent tracking parameters throughout testing. ByteTrack was configured with a detection confidence threshold of 0.5, a track buffer size of 10 frames, and an IoU matching threshold of 0.8. BOT-SORT used these same core parameters while incorporating additional high and low confidence thresholds of 0.6 and 0.1, respectively. These configurations were based on the trackers' original implementations and optimized for our specific tracking scenario while preserving their fundamental operational characteristics.

\subsubsection{Results}
After fine-tuning YOLOX-L on our dataset, we evaluated both the fine-tuned and off-the-shelf models as detection backends for tracking. Table \ref{tab:detection-eval} presents the per-class detection results. All evaluations used a detection confidence threshold of 0.4 and an NMS threshold of 0.5. While NMS is typically set higher, the densely packed objects and frequent bounding box overlaps in our dataset required a lower threshold to prevent excessive suppression. The fine-tuned YOLOX-L model significantly improved recall across all object classes, leading to higher F1 scores, particularly for pedestrians, motorbikes, and cyclists. The F1 score increased by approximately 2.5x across classes compared to the off-the-shelf model. While for vehicles, the precision dropped from 90.14 to 85.53, this trade-off resulted in a substantial reduction in false negatives (FN), reducing overall missed detections from 85,447 to 36,469. This improvement is crucial for tracking, where recall is essential for maintaining track consistency. Error analysis highlights a key limitation of the off-the-shelf model. Its high precision comes at the cost of poor recall, leading to frequent missed detections. This performance gap is particularly pronounced in challenging scenarios, such as high object density and adverse conditions like rainy nights. Fine-tuning on the train split enabled better detection robustness, especially for underrepresented classes.

Table \ref{tab:tracking-eval} presents tracking results across three detection settings and two tracking models, yielding six result combinations. BoT-SORT consistently outperforms ByteTrack in HOTA and MOTA across all settings, except for motorbike HOTA and pedestrian MOTA in the off-the-shelf setting. The largest performance gap is observed in the ground-truth setting for cyclists, where BoT-SORT achieves a MOTA improvement of 20 and HOTA improvement of 16.7 over ByteTrack. The low false negatives (FN) and high false positives (FP) in the off-the-shelf setting stem from the detector’s low recall, making it overly permissive and generating excessive detections with poor precision. For instance, in ByteTrack, FP for vehicles drops from 124,039 (ground-truth) to 20,924 (off-the-shelf), but FN increases more than 2.5×, from 162,065 to 441,511, confirming that the model struggles with false detections despite detecting more objects. 

This imbalance impacts ID assignment, as seen in BoT-SORT’s pedestrian IDs, which drop from 960 (fine-tuned) to 55 (off-the-shelf), indicating that the tracker aggressively associates detections by leveraging high recall. However, this comes at the cost of tracking false objects and ghost tracklets, leading to inconsistent ID management. The low IDF1 scores in BoT-SORT (68.2 and 69.9 in fine-tuned vs. 5.9 and 6.4 in off-the-shelf) further confirm this issue, as the tracker frequently initializes new tracklets instead of maintaining consistent tracking IDs. While the fine-tuned setting significantly improves over the off-the-shelf model, the performance gap between fine-tuned and ground-truth settings underscores tracking sensitivity to detector errors. For instance, in BoT-SORT, pedestrian MOTA drops from 84.9 (ground truth) to 58.7 (fine-tuned), and motorbike IDF1 declines from 76.8 to 58.1, showing that fine-tuning is beneficial but does not fully close the gap. These results suggest that further improvements in both detection and tracking are needed to enhance overall tracking stability, reduce fragmentation, and maintain consistent object identities.

\begin{table*}[t]
\centering
\caption{Trajectory Predictors Evaluation (Sequential)}
\resizebox{\textwidth}{!}{
\begin{tabular}{|c|p{3cm}|cc|cc|cc|cc|cc|cc|cc|}
\hline

& \multirow{2}{*}{\textbf{Predictor}} & \multicolumn{2}{|c|}{\textbf{10/10, 1s/1s}} &  \multicolumn{2}{|c|}{\textbf{20/10, 2s/1s}} &  \multicolumn{2}{|c|}{\textbf{10/20, 1s/2s}} &  \multicolumn{2}{|c|}{\textbf{20/20, 2s/2s}} &  \multicolumn{2}{|c|}{\textbf{20/30, 2s/3s}}  &  \multicolumn{2}{|c|}{\textbf{20/60, 2s/6s}} \\ \cline{3-14} 
& & \textbf{ADE↓} & \textbf{FDE↓} & \textbf{ADE↓} & \textbf{FDE↓} & \textbf{ADE↓} & \textbf{FDE↓} & \textbf{ADE↓} & \textbf{FDE↓} & \textbf{ADE↓} & \textbf{FDE↓} & \textbf{ADE↓} & \textbf{FDE↓}  \\ \hline

\multirow{3}{*}{\rotatebox{90}{\textbf{\makecell{sliding\\window=1}}}} & \textit{LSTM} & \textbf{10.49} & \textbf{19.63} & 9.52 & 17.62 & 19.20 & 42.43 & 17.46 & 38.21 &  25.86 & 59.84 & 46.27 &  105.68\\ 
& \textit{Transformer} & 10.66 & 20.49 & 9.23 & 17.29 & 19.48 & 43.05  &  18.45 & 40.70 & 27.86 & 64.40 &  46.88 &  105.47 \\ 
& \textit{Transformer+GMM} & 10.76 & 20.22 & 9.68 & 17.88 & 19.48 & \textbf{41.71}  & 18.36 & 39.14&26.36& 57.48 &\textbf{44.94} & 96.89 \\ \hline

\multirow{3}{*}{\rotatebox{90}{\textbf{\makecell{sliding\\window=3}}}} &  \textit{LSTM} & 10.55 & 19.87 &  9.50 & 17.40 & \textbf{19.18} & 42.10 & 17.08 & 37.09 & 24.94 &  57.58 & 45.81 & 103.67\\  
&  \textit{Transformer}  &11.13&21.49 &10.19  &19.24  &20.66 &47.53  &18.49  &40.53  &28.21 &64.63  & 46.06 & 101.31 \\ 
&  \textit{Transformer+GMM} & 10.77 & 20.24 & 9.74 & 17.97 & 19.80 & 41.99 & 17.61 & 37.50 & 26.86 & 57.12 & 45.46 & \textbf{93.28} \\   \hline

\multirow{3}{*}{\rotatebox{90}{\textbf{\makecell{sliding\\window=5}}}} &  \textit{LSTM} & 10.77 & 20.12 & \textbf{9.20} &  \textbf{16.84} & 19.54 & 43.10 & \textbf{17.04}  & 37.02 &  \textbf{24.91} & 57.06 & 45.92 & 104.84  \\ 
&  \textit{Transformer}  & 11.30 & 21.71  & 9.72  & 18.47 & 21.92  & 52.03  & 19.86  & 45.57 & 27.64 & 63.30  & 47.31 & 105.54  \\  %23.27 43.92 -> seed 42 -> 24.78 48.42 
&  \textit{Transformer+GMM} & 10.99 & 20.55 & 9.85 & 18.02 & 20.05 & 42.68 & 17.40 & \textbf{36.47} & 25.79 & \textbf{56.54} &  45.67& 93.95  \\   \hline
\end{tabular}
}
\label{tab:prediction_results}
\end{table*}

\begin{table*}[t]
\centering
\caption{Trajectory Predictors Evaluation (Multimodal, Sequential)}
\resizebox{\textwidth}{!}{
\begin{tabular}{|c|p{3cm}|cc|cc|cc|cc|cc|}
\hline

& \multirow{2}{*}{\textbf{Predictor}} & \multicolumn{2}{|c|}{\textbf{10/10, 1s/1s}} &  \multicolumn{2}{|c|}{\textbf{10/20, 1s/2s}} &  \multicolumn{2}{|c|}{\textbf{20/30, 2s/3s}}  &  \multicolumn{2}{|c|}{\textbf{20/60, 2s/6s}} \\ \cline{3-10} 
& & \textbf{MinADE$_5$↓} & \textbf{MinFDE$_5$↓} & \textbf{MinADE$_5$↓} & \textbf{MinFDE$_5$↓} & \textbf{MinADE$_5$↓} & \textbf{MinFDE$_5$↓} & \textbf{MinADE$_5$↓} & \textbf{MinFDE$_5$↓}  \\ \hline

\textbf{\makecell{sliding\\window=1}} & \textit{Transformer+GMM} & \textbf{7.19} & \textbf{12.01} & \textbf{13.11} & \textbf{24.75}  & 19.93 & 41.42  & \textbf{32.30} & \textbf{62.23}   \\ \hline
\textbf{\makecell{sliding\\window=3}} & \textit{Transformer+GMM}  &  7.27&12.19  &14.11  &27.82 & 20.42& 40.44  &35.61 &67.69  \\  \hline
\textbf{\makecell{sliding\\window=5}} &  \textit{Transformer+GMM} &7.49 &12.72 &13.19 &25.56 & \textbf{17.32} & \textbf{34.24} &35.49 &66.41  \\  \hline
\end{tabular}
}
\label{tab:prediction_results_multimodal}
\end{table*}

\begin{table*}[t]
\centering
\caption{Trajectory Predictors Evaluation (Frame-based)}
\resizebox{0.8\textwidth}{!}{%
\begin{tabular}{|c|p{3cm}|cc|cc|cc|cc|cc|}
\hline
& \multirow{2}{*}{\textbf{Predictor}} & \multicolumn{2}{|c|}{\textbf{video\_122233 }} &  \multicolumn{2}{|c|}{\textbf{video\_155425}} &  \multicolumn{2}{|c|}{\textbf{video\_160325}} &  \multicolumn{2}{|c|}{\textbf{video\_125233}} &  \multicolumn{2}{|c|}{\textbf{video\_204347}}  \\ \cline{3-12} 
& & \textbf{ADE↓} & \textbf{FDE↓} & \textbf{ADE↓} & \textbf{FDE↓} & \textbf{ADE↓} & \textbf{FDE↓} & \textbf{ADE↓} & \textbf{FDE↓} & \textbf{ADE↓} & \textbf{FDE↓} \\ \hline
\multirow{6}{*}{\rotatebox{90}{\textbf{prediction horizon=10}}} 
 & \textit{LSTM} & \textbf{7.55} & \textbf{13.94} &  \textbf{30.92} &  \textbf{56.57} & 25.16  & 44.83  & \textbf{9.04}  & \textbf{17.1}  & \textbf{17.73}   & \textbf{34.73} \\ 
 & \textit{Transformer}  &  8.28 & 15.77 &  31.88 & 58.12  & \textbf{24.68} & \textbf{44.48} & 9.71 & 18.69 & 23.28 & 47.82 \\
 & \textit{GCN} & 9.57 & 16.18 & 31.92 & 56.93 & 27.62 & 47.95 & 11.7 &  21.28  &  24.51 & 46.13 \\
 & \textit{GCN$_{temporal}$}  & 10.2 & 19.1 & 31.62 & 58.72 & 26.32 & 46.58 & 12.5 &  23.64 &  23.88 & 47.30  \\ 
 & \textit{GAT} & 10.13 & 17.91 & 33.18 & 59.62 & 26.98 & 47.32 & 11.98 & 22.02 & 25.28 & 49.46  \\
 & \textit{GAT$_{temporal}$}  & 10.54 & 19.77 & 32.11  & 59.43 & 26.14 & 47.16 & 13.1 & 24.75 & 24.76 & 49.08  \\ \hline
\multirow{6}{*}{\rotatebox{90}{\textbf{prediction horizon=20}}} 
 & \textit{LSTM} &  \textbf{14.5} & \textbf{32.63} &  60.83 & 125.53 & 47.15 & 96.83  & \textbf{18.18}  & \textbf{40.69} &  \textbf{35.99}  & \textbf{85.21} \\ 
 & \textit{Transformer}  & 17.3 & 40.5 & 65.59 & 132.95 & 50.44  & 103.07 & 19.46 & 43.43 &  48.51 & 116.02 \\
 & \textit{GCN} & 17.94 & 35.22 & 66.05 & 132.18 &  53.9 & 102.78  & 23.74  & 47.89  &  49.69 & 108.02  \\
 & \textit{GCN$_{temporal}$}  & 18.88  & 41.38 & 65.55 & 133.08 & 50.18 & 101.55 & 24.07 & 51.98 & 46.13 & 105.42 \\ 
 & \textit{GAT} & 17.17 & 38.33 & \textbf{58.6} & \textbf{119.8} & \textbf{44.29} & \textbf{88.44} & 21.77  & 46.03 & 48.2 & 112.45\\
 & \textit{GAT$_{temporal}$}  & 19.62 & 42.94 & 65.27 & 133.09 & 51.57& 104.84 & 24.29 &  53.46 & 47.31 & 109.49  \\ \hline
\end{tabular}%
}
\label{tab:prediction_results_frame}
\end{table*}

\subsection{Trajectory Prediction}
\label{sec:Prediction}

For trajectory prediction, three deep-learning architectures are evaluated, each differing in their ability to capture temporal dependencies and interaction dynamics. 

\subsubsection{Predictors}
\begin{itemize}
    \item \textbf{LSTM-based models} assume that motion follows temporally causal dependencies, making recurrent networks a common choice for trajectory prediction. By leveraging hidden states to incorporate past information, LSTMs capture dynamic motion properties and temporal correlations in prediction sequences.
    
    \item \textbf{Transformer-based models} employ self-attention layers within an encoder-decoder architecture to capture dependencies comprehensively. The encoder processes input trajectories with positional encodings to preserve temporal order, while self-attention mechanisms compute pairwise relationships across timesteps. We adopt the Transformer configuration proposed by \cite{Murad}, which integrates encoder-decoder Transformer layers to model sequential dependencies from observed trajectories, combined with a Mixture Density Network (MDN) to estimate Gaussian Mixture Model (GMM) parameters. For generating a deterministic unimodal trajectory, the mean ($\mu$) of the mixture component with the highest mixing coefficient ($\pi$) at each timestep is selected. For multimodal predictions, trajectory points can either be sampled from the mixtures or taken directly as their mean values ($\mu$), considering only mixtures with mixing coefficients $\pi > \tau$. The trajectory module then generates paths by connecting these points across time steps, where each point at time $t+1$ is connected to its closest neighbor at time $t$.
    
    \item \textbf{Graph Neural Networks (GNNs)} model complex spatial and temporal interactions by operating on graphs constructed from each scene, where nodes represent traffic agents and edges encode pairwise relationships, such as proximity. Each node contains information about an agent's past trajectory. We evaluate two message-passing strategies: \textbf{Graph Convolutional Networks (GCNs)}, which uniformly aggregate information from neighboring nodes, and \textbf{Graph Attention Networks (GATs)}, which dynamically compute attention weights to prioritize more influential interactions. Both GNNs are tested in two configurations: (1) \textit{sequence-to-sequence}, where the past trajectory is flattened and directly used for future trajectory prediction, and (2) \textit{temporal}, where a GNN encoder first captures spatial interactions, followed by an LSTM encoder-decoder to model temporal dynamics and generate future trajectories in an autoregressive manner.
  
\end{itemize}

\begin{figure*}[h!]
\centering
\includegraphics[width=1.0\textwidth]{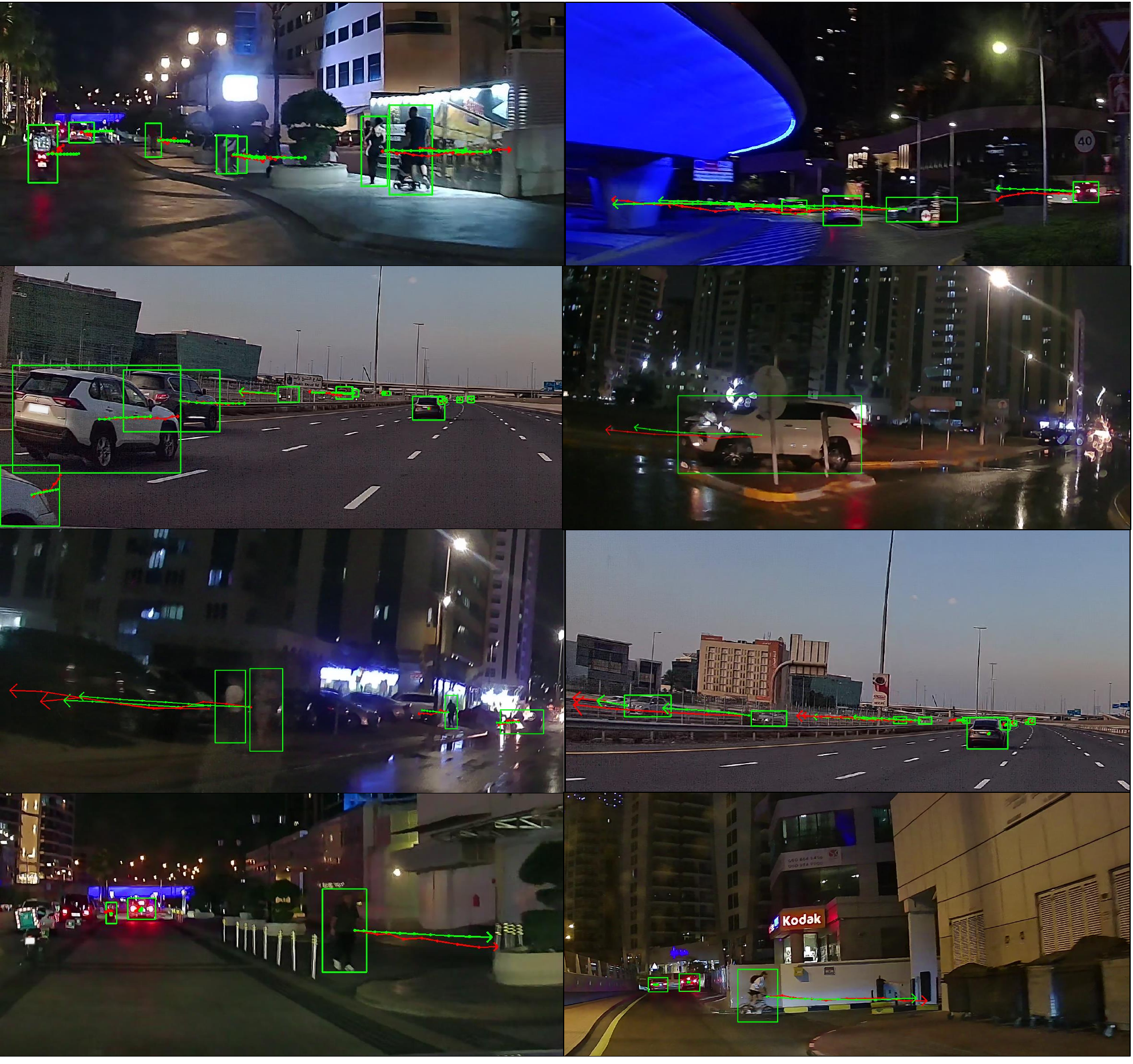}
\caption{Samples of predicted trajectories: ground truth trajectories are shown in red, while predictions are depicted in green. The predicted trajectories accurately capture the agent's heading but exhibit lower accuracy in predicting speed, which in turn affects the precise estimation of future trajectory locations.}
\label{fig:prediction_samples}
\end{figure*}

\subsubsection{Evaluation Protocol}
Each predictor is trained on ground truth past trajectories. To compare performance, we compute the average and final displacement errors for each prediction setting in a unimodal mode, as well as for the Transformer coupled with the GMM model in a multimodal output setting. Let $\zeta_{i,t}$ and $\hat{\zeta}_{i,t}$ denote the predicted and ground truth trajectories for the $i$-th agent at time step $t$, respectively.

\begin{itemize}
  \item \textit{Average Displacement Error (ADE)} measures the average $\ell_2$ distance between predicted and ground truth trajectories over the entire prediction horizon $\Delta t$:
    \begin{equation*}
        ADE = \frac{1}{n \times \Delta t} \sum_{i=1}^{n} \sum_{t=1}^{\Delta t} ||\zeta_{i,t} - \hat{\zeta}_{i,t}||
    \end{equation*}

    For multimodal predictions:
    \begin{equation*}
        MinADE_k = \frac{1}{n \times \Delta t} \min_{k} \sum_{i=1}^{n} \sum_{t=1}^{\Delta t} ||\zeta^k_{i,t} - \hat{\zeta}_{i,t}||
    \end{equation*}

  \item \textit{Final Displacement Error (FDE)} measures the average $\ell_2$ distance between predicted and ground truth trajectories at the final time step $\Delta t$:
    \begin{equation*}
        FDE = \frac{1}{n} \sum_{i=1}^{n} ||\zeta_{i, \Delta t} - \hat{\zeta}_{i, \Delta t}||
    \end{equation*}

    For multimodal predictions:
    \begin{equation*}
        MinFDE_k = \frac{1}{n} \min_{k} \sum_{i=1}^{n} ||\zeta^k_{i, \Delta t} - \hat{\zeta}_{i, \Delta t}||
    \end{equation*}
\end{itemize}

For multimodal evaluation, the metric computes the $\ell_2$ distance between the ground truth trajectory and the closest prediction among $k$ possible trajectories, with $k$ set to 5.
\subsubsection{Implementation Details}
All predictors operate on normalized relative distances, computed as differences between absolute positions and standardized using the mean and standard deviation from the training dataset. This normalization strategy mitigates the impact of varying resolutions and improves training convergence. Both LSTM and Transformer models process sequences in the format \([past\_trajectory]\ [future\_trajectory]\).

The LSTM implementation uses two layers with a hidden dimension of 128, trained with a batch size of 128 for 50 epochs. Our transformer architecture features three encoder-decoder layers, each employing four attention heads and an embedding dimension of 128. To maintain temporal causality during training, the decoder employs a masked attention mechanism on shifted target trajectories, while the encoder processes the past trajectory input. At inference time, both models autoregressively generate predictions using outputs from previous time steps, maintaining the normalized relative position representation. All models are trained using Mean Squared Error (MSE) loss. For the transformer, we adapt the learning rate scheduler parameters and warmup period to accommodate significantly different dataset sizes resulting from various sliding window configurations. Gradient clipping ensures stable training across these configurations. The final predicted trajectories are denormalized to absolute positions before computing the ADE and FDE metrics.

The transformer-GMM implementation extends the standard transformer with a specialized Gaussian Mixture Model component for probabilistic trajectory prediction. The architecture features three encoder-decoder layers with four attention heads and an embedding dimension of 128. The GMM layer is implemented through three parallel networks, each structured with three fully-connected layers using ELU activations and progressively decreasing widths (from model dimension to a hidden size of 32, then to 24, and finally to 16 units). These networks specialize in generating different GMM parameters: mixture weights ($\pi$), means ($\mu$), and covariance matrices ($\sigma$) for six two-dimensional Gaussians. During training, a masked attention mechanism in the decoder is employed while passing zeros (zero conditioning) as inputs rather than shifted targets. The model is trained for 100 epochs using negative log-likelihood (NLL) loss, with adaptive learning rate scheduling and gradient clipping to ensure stability. At inference time, the model produces complete future trajectories in a single forward pass, with the GMM outputs fed into a future trajectory building algorithm that prioritizes high-probability components while maintaining alternative paths.

For GNN models, the input is reformatted into a frame-centered structure. For each frame, the data is rearranged such that all objects contained within the frame are aggregated along with their past and future trajectories. If the available past or future trajectories are shorter than the predefined observation length and future horizon, the corresponding object is disregarded for that frame. To feed frame-based data into GNNs, the parameter \texttt{max\_num\_nodes} is set to 50. This is necessary since PyTorch's GNN implementation operates with a static adjacency matrix. To handle the varying number of agents across frames, masking is implemented. Each sample consists of a fixed number of nodes representing agents in a scene and includes past trajectory features, an adjacency matrix, future trajectory targets, and a mask indicating valid nodes.

To construct the mask, given a batch with a variable number of agents \( n \) (up to \texttt{max\_num\_nodes}), the mask is computed as $M_i =1, \text{if } i \leq n$ and $0, \text{otherwise}$. This mask is applied during loss computation to prevent gradient updates on padded entries. Edges in the adjacency matrix are established if the pairwise distance between nodes is below a predefined threshold, which is set to 100. To ensure stable training and effective message passing, the adjacency matrix is symmetrically normalized:

\[
\hat{A} = D^{-\frac{1}{2}} (A + I) D^{-\frac{1}{2}},
\]

where \( D \) is the degree matrix. This normalization prevents numerical instability caused by scale differences between nodes and ensures effective information propagation across the graph. The sequence-to-sequence models consist of two GNN encoder and decoder layers with a hidden dimension of 256, trained with a batch size of 32 for 50 epochs. The temporal variation of the models consists of two GNN encoder layers with a hidden dimension of 256, two LSTM encoders and decoders  with a hidden dimension of 2128. A masked Mean Squared Error (MSE) loss is used to ensure that missing or padded nodes do not contribute to the loss function:

\[
\mathcal{L} = \frac{1}{\sum_{b,n} M_{b,n}} \sum_{b,t,n} M_{b,n} \| \hat{V}_{b,t,n} - V_{b,t,n} \|^2,
\]

where \( M_{b,n} \) is a mask indicating valid nodes, \( \hat{V}_{b,t,n} \) is the predicted velocity, and \( V_{b,t,n} \) is the ground-truth velocity.

\subsubsection{Results}

Table \ref{tab:prediction_results} presents the evaluation results of LSTM, Transformer, and Transformer coupled with GMM on prediction datasets with varying prediction horizons, observed trajectory lengths, and sliding window settings. The best results for each column are highlighted in bold, indicating which model and sliding window configuration performed best. For longer prediction horizons (i.e., 20, 30, and 60), the Transformer coupled with GMM outperforms the other models in terms of final displacement error (FDE). In contrast, for an observed trajectory of 10 and a prediction horizon of 1, LSTM achieves the best performance. The first two settings in the table (10/10 and 20/10) test the hypothesis that a longer observed trajectory improves prediction accuracy. This hypothesis is supported by the results, as using an observed trajectory of 20 achieves an ADE of 9.20 compared to 10.49 for an observed trajectory of 10. Similarly, the FDE improves from 19.63 to 16.84. A similar trend is observed in the third and fourth columns for a prediction horizon of 20 with observed trajectories of 10 and 20. The best settings demonstrate an FDE improvement of 5.36 and an ADE improvement of 2.04 when increasing the observed trajectory length from 10 to 20.

Table \ref{tab:prediction_results_multimodal} presents the results of the multimodal output evaluation, with the best results highlighted in bold across different sliding window settings. The model performed best when trained using a sliding window of 1. Compared to the unimodal setting, the multimodal approach demonstrates a clear advantage. For instance, in the 20/60 split, the best ADE in the unimodal setting is 44.94, whereas the multimodal model achieves 32.30. Similarly, FDE improves from 93.28 (unimodal) to 62.23 (multimodal), representing a reduction of over 30. In the simplest setting (10/10), the multimodal model outperforms LSTM, achieving a gain of 3.3 in ADE and 7.6 in FDE. Further, Table \ref{tab:prediction_results_frame} presents the results of frame-based evaluation for prediction horizons of 10 and 20. To better reflect real-world applications, objects without a sufficiently long past trajectory, matching the model's required observation length, are padded with zeros to simulate newly detected objects entering the scene. For a prediction horizon of 10, LSTM demonstrates the most robust performance across all test videos, except for the third sample, where the Transformer achieves a better score by 0.48 in ADE and 0.35 in FDE. In the case of a longer prediction horizon, GAT outperforms LSTM in the second and third samples, achieving more tangible improvements, with an average gain of 2.5 in ADE and 7.06 in FDE. Figure \ref{fig:prediction_samples} illustrates examples of predicted trajectories, where LSTM-generated predictions in the 10/10 setting are shown with green arrows, while ground truth trajectories are depicted in red. The predictions accurately capture the agent's heading but show lower accuracy in precise location estimation, particularly struggling to adapt to varying agent speeds.

\subsection{Intention Prediction}

%%%%%%%%%%%%%%%%%%%%%%%%%%%%%%%%%%%%%%%%%%%%%%%%%%%%%%
%%% Intention Evaluation

\begin{table*}[t]
\centering
\caption{Intention Prediction Evaluation (I - vanilla setting and II - autoregressive mode).}
\resizebox{\textwidth}{!}{
\begin{tabular}{|p{1.3cm}|p{2cm}|c|c|c|c|c|c|}
\hline
\multirow{2}{*}{\makecell{\textbf{Prediction} \\ \textbf{Setting}}} & \multirow{2}{*}{\textbf{Intention}} & \multicolumn{2}{|c|}{\textbf{10/10, 1s/1s}} &  \multicolumn{2}{|c|}{\textbf{15/10, 1.5s/1s}} &  \multicolumn{2}{|c|}{\textbf{20/10, 2s/1s}}  \\ \cline{3-8}
 & & \textbf{F1-score$_{all}$(\%)↑} & \textbf{F1-score$_{last}$(\%)↑} & \textbf{F1-score$_{all}$(\%)↑} & \textbf{F1-score$_{last}$(\%)↑} & \textbf{F1-score$_{all}$(\%)↑} & \textbf{F1-score$_{last}$(\%)↑}  \\ \hline
 
\multirow{11}{*}{\textit{LSTM}$_{I}$} & reversing & 4.73 & 4.98 & 54.72 & 50.27 & \textbf{75.76} & \textbf{78.22} \\ 
& turn-right & 22.28 & 22.48 & 57.94 & 56.66 & \textbf{78.33} & \textbf{76.76} \\ 
& turn-left & 17.22 &  17.17 & 57.99 & 56.79 & \textbf{76.56} & \textbf{73.36}  \\ 
& merge-right & 11.57 & 11.31 & 52.95 & 51.54 & \textbf{76.07} & \textbf{74.41} \\ 
& merge-left & 20.37 &  19.07 & 58.46 & 56.54& \textbf{78.81} & \textbf{76.63} \\   
& braking & 1.66 & 1.76 & 20.39 & 19.11 & \textbf{34.75} & \textbf{34.49} \\ 
& stopped & 68.66 & 68.18 & 84.08 & 83.61 & \textbf{90.62} & \textbf{90.07} \\ 
& lane-keeping & 89.99 & 89.92 & 94.74 & 94.61 & \textbf{96.95} & \textbf{96.77}  \\
& walking & 32.82 &  33.33 & 70.74 & 70.86 & \textbf{86.44} & \textbf{86.48}  \\
& crossing & 38.59 & 37.43 & 69.06 & 67.29 & \textbf{85.96} & \textbf{85.74} \\ 
& waiting\_to\_cross & 24.29 &  23.41 & 63.87 & 63.3 & \textbf{83.55} & \textbf{83.24}  \\ \cline{2-8}
& \textbf{overall score} & 30.2 & 29.91 & 62.27 & 60.96 & \textbf{78.53} & \textbf{77.83} \\ \cline{2-8}
& \textbf{$D_{L_{norm}}$} & \multicolumn{2}{|c|}{0.1708} &  \multicolumn{2}{|c|}{0.09} & \multicolumn{2}{|c|}{\textbf{0.0517}} \\ \hline

\multirow{11}{*}{\textit{LSTM}$_{II}$}  & reversing  & 12.70 & 10.53 & 50.75 & 47.61 & 72.82 & 73.31 \\ 
& turn-right & 18.23 & 17.91 & 59.33 & 58.67 & 75.62 & 73.58  \\ 
& turn-left & 15.65 & 15.12 & 56.29 & 54.92 & 73.08 & 72.26 \\ 
& merge-right & 12.58 & 12.37 & 52.66 & 51.60 & 73.67 & 72.27 \\ 
& merge-left & 20.21 & 19.46 & 57.92 & 56.73 & 77.01& 74.77 \\ 
& braking & 2.99 & 3.36 & 16.05 & 16.22 & 33.48 & 32.53 \\
& stopped & 68.47 & 67.92 & 83.55 & 83.06 & 90.26 & 89.84\\ 
& lane-keeping & 89.93 & 89.85 & 94.66 & 94.55 & 96.83& 96.69\\
& walking & 31.93 & 32.51 & 69.44 & 69.91 & 83.76 & 83.92 \\ 
& crossing & 36.03 & 35.16 & 69.88 & 68.42 & 82.55 & 81.86 \\ 
& waiting\_to\_cross & 23.91 & 23.83 & 61.29 & 60.55 & 79.01 & 80.03\\ \cline{2-8}
& \textbf{overall score} & 30.24 & 29.82 & 61.08 & 60.20 & 76.19 & 75.55 \\ \cline{2-8}
& \textbf{$D_{L_{norm}}$} & \multicolumn{2}{|c|}{0.1716} & \multicolumn{2}{|c|}{0.0917} & \multicolumn{2}{|c|}{0.0544} \\ \hline
\end{tabular}
}
\label{tab:intention_results}
\end{table*}
 %%%%%%%%%%%%%%%%%%%%%%%%%%%%%%%%%%%%%%%%%%%%%%%%%%%%%%%%%%%%%%%%

\begin{figure*}[t!]
\centering
\includegraphics[width=1.0\textwidth]{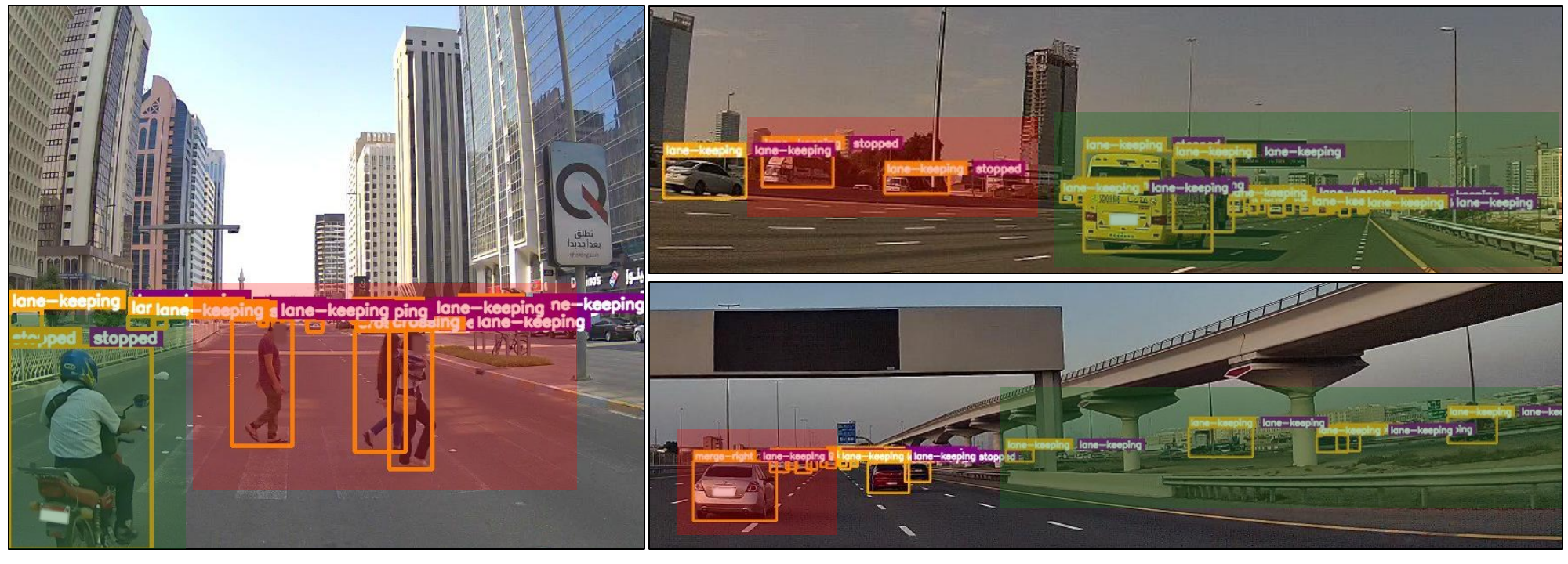}
\caption{Samples of predicted intentions during daytime. The ground truth intention for the next timestamp is shown in orange, while predictions are in purple. Red rectangles highlight misclassifications, such as predicting "keep\_lane" for walking and crossing, failing to predict "merge," and incorrectly assigning "stop" to a reversing vehicle. Green indicates correctly predicted intentions, including accurate "keep\_lane" predictions on the main road, successful "merge" predictions from the right road, and correct stopping behavior at crossings.}
\label{fig:intention_day}
\end{figure*}

\begin{figure*}[h!]
\centering
\includegraphics[width=1.0\textwidth]{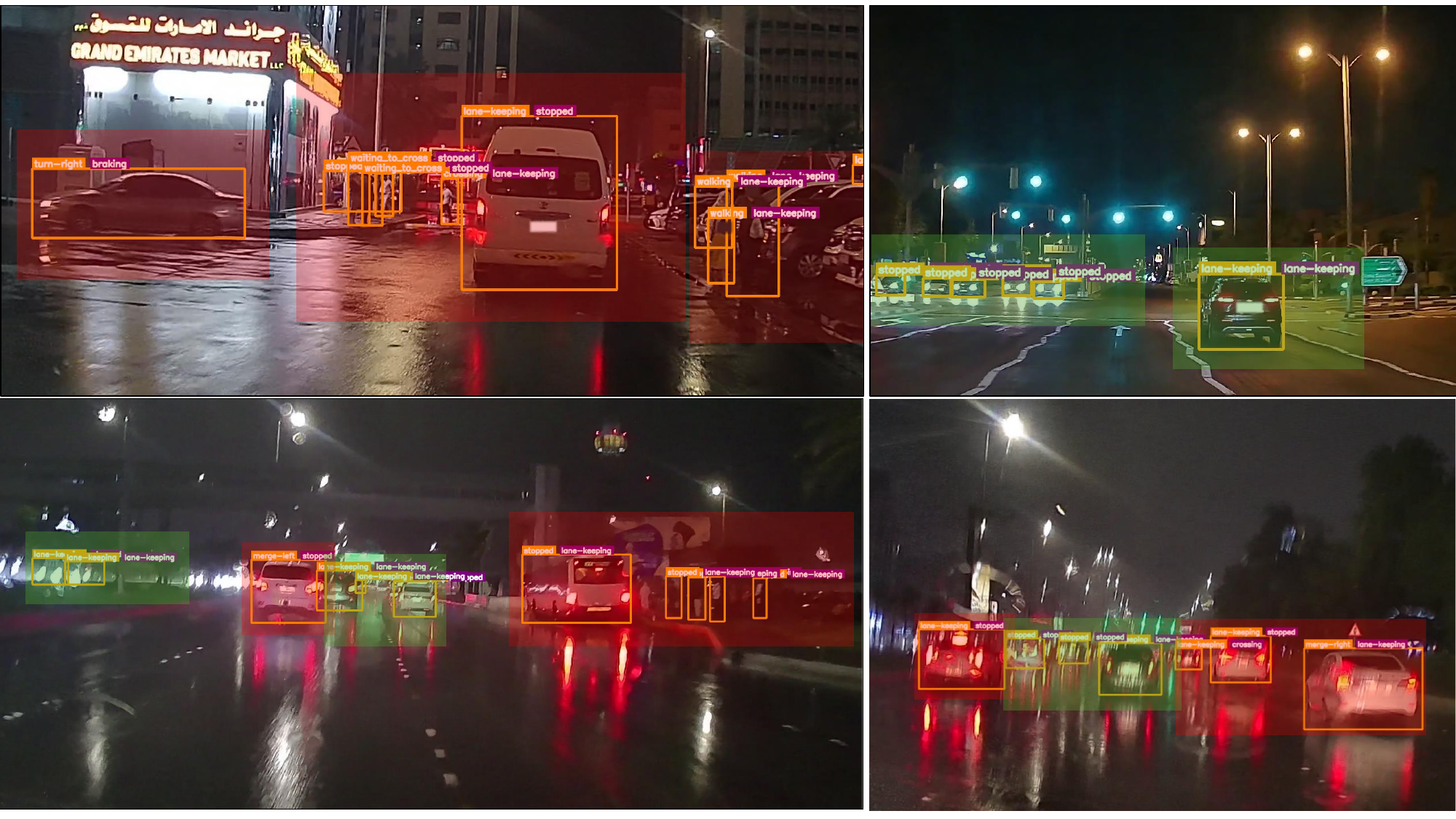}
\caption{Samples of predicted intentions during nighttime. The ground truth intention for the next timestamp is shown in orange, while predictions are in purple. Red rectangles highlight misclassifications, such as predicting "stop" for moving vehicles and failing to predict "merge" and "turn." Green indicates correctly predicted intentions, including accurate stopping behavior of vehicles at intersections.}
\label{fig:intention_night}
\end{figure*}

For intention prediction, the experiments involve providing the model with a past trajectory and generating future intentions at each timestamp over a prediction horizon of 10. The core evaluation model is an LSTM trained in a manner to classify and predict the most probable intention from a set of 11 classes. The model is evaluated in two settings: \textit{vanilla}, where predicted intentions are assumed to be independent, and \textit{autoregressive}, where dependencies between consecutive predictions are considered. For instance, if a vehicle has stopped at a junction, it is less likely to change lanes immediately upon resuming movement; instead, it is more probable that it will continue with a lane-keeping intention. Similarly, for pedestrians, a "waiting to cross" intention is likely to be followed by "crossing".

\subsubsection{Predictors}
\begin{itemize}
    \item \textbf{LSTM$_{I}$ model} corresponds to the vanilla setting, where the decoder generates the entire sequence of future intentions in a single forward pass. Predictions are evaluated against the ground truth using CrossEntropyLoss.
    \item \textbf{LSTM$_{II}$ model} operates in an autoregressive mode with the capability of teacher forcing during training. The decoder predicts intentions one timestep at a time, appending the previous output as input for the next step. During inference, the model relies entirely on its own predictions, with teacher forcing set to zero.
\end{itemize}

\subsubsection{Evaluation Protocol}
Intention prediction is formulated as a classification task, where the intention is predicted over a 10-timestep horizon and the evaluation is conducted using 5-fold cross-validation. To assess the impact of the observed trajectory length on prediction accuracy, we evaluate the model under three settings, each with a fixed future horizon of 10 frames: using the past \textit{10}, \textit{15}, and \textit{20} frames as input. The models' performance is evaluated using the following metrics:

\begin{itemize}
    \item \textit{F1-score for all timestamps:} This score is computed over every token in the predicted sequence by flattening all timesteps. The F1-scores are computed both per class and overall using macro averaging, which treats each class equally. This ensures that the performance on underrepresented classes is adequately reflected.

    \item \textit{F1-score for the last token:} In addition to the overall F1-score, we evaluate the robustness of the prediction by computing the F1-score for the last token (i.e., the final intention) of the sequence. This score is computed by comparing the final token in  predicted intention sequence to the corresponding token in the ground truth intention sequence, with macro averaging applied in the same manner as for the overall F1-score.

    \item \textit{Average Normalized Levenshtein Distance \cite{4160958}:} To assess the performance at the sequence level, we compute the Levenshtein distance between the ground truth sequence (\( s^{gt} \)) and the predicted sequence (\( s^{pr} \)) . The Levenshtein distance, denoted \( D_L \), is defined as the minimum number of single-token edits (insertions, deletions, or substitutions) required to transform the grounf truth sequence to predicted. A normalized Levenshtein Distance ($D_{L_{\text{norm}}}$), when equals 0 indicates a perfect match, while values closer to 1 mean larger discrepancies. To compute the distance, let \( D(i, j) \) denote the Levenshtein distance between the first \( i \) tokens of \( s^{gt} \) and the first \( j \) tokens of \( s^{pr} \). The recursive formulation is given by
    \begin{equation*}
        \resizebox{0.93\columnwidth}{!}{$
        D(i, j) =
        \begin{cases}
            \max(i, j), & \text{if } \min(i, j) = 0, \\[8pt]
            \min \begin{cases}
                D(i-1, j) + 1, \\
                D(i, j-1) + 1, \\
                D(i-1, j-1) + \delta(s^{gt}_i, s^{pr}_j)
            \end{cases}, & \text{otherwise}.
        \end{cases}
            $}
    \end{equation*}, where \( s^{gt}_i \) as the \( i \)-th token of the ground truth sequence, \( s^{pr}_j \) is the \( j \)-th token of the predicted sequence and $\delta$ is defined as:  
    \begin{equation*}
        \delta(s^{gt}_i, s^{pr}_j) =
        \begin{cases}
            0, & \text{if } s^{gt}_i = s^{pr}_j \\
            1, & \text{if } s^{gt}_i \neq s^{pr}_j
        \end{cases}
    \end{equation*}
    
    The computed distance is then normalized by dividing by the length of the ground truth sequence \( |s^{gt}| \):
    \begin{equation*}
        D_{L_{\text{norm}}}(s^{gt}, s^{pr}) = \frac{D_L(s^{gt}, s^{pr})}{|s^{gt}|}.
    \end{equation*}
\end{itemize}

\subsubsection{Implementation Details}
Both models were trained with a batch size of 128, using two hidden layers of size 128, over 50 epochs. Two evaluation settings are implemented: (1) a train/test split, similar to that used in trajectory prediction, and (2) \(k\)-fold cross-validation. Cross-validation was conducted to provide a comprehensive assessment of model performance, given the diversity of the dataset and the fact that the models rely solely on positional data. For a more thorough evaluation and testing generalization of prediction models, especially when transferring models trained on other intention datasets to EMT, the train/test split option is also provided and recommended. Additionally, for the LSTM autoregressive mode, a teacher forcing ratio can be set, allowing for experimentation with the degree of ground truth feedback introduced during training.

\subsubsection{Results}
Table~\ref{tab:intention_results} presents the results of the conducted experiments, detailing the performance across both model settings, intention classes, and past trajectory settings. The highest performance is achieved by the vanilla model configuration, surpassing the autoregressive setting by nearly 2\% in F1-score across all timestamps. The results indicate that when only relative displacement is utilized for prediction, the model fails to capture temporal dependencies between intentions. Intentions such as "waiting\_to\_cross" and "crossing" should exhibit sequential dependencies. However, as shown in Figure~\ref{fig:intention_day}, the model incorrectly predicts the "keep\_lane" intention for crossing scenarios. This misclassification occurs due to the similarity between lane-keeping motion patterns and the vehicle-relative displacement of crossing pedestrians. Additionally, since the model does not incorporate object type information or associated intention priors, it is unable to distinguish between different entities. 

A similar issue is observed for "waiting\_to\_cross," as shown in Figure~\ref{fig:intention_night}, where the model does not differentiate between a pedestrian standing at a crossing and one waiting at a bus stop. This limitation arises from the absence of visual cues in the feature set. Additionally, since the model primarily relies on trajectory patterns, an object may appear to be moving due to shifts in its bounding box caused by the camera perspective, despite being stationary in the real world. As a result, the model incorrectly classifies the object as dynamic. A clear trend observed in the results is the significant improvement in performance with increasing past trajectory length. The most frequent intention, "keep\_lane," benefits from an improvement of approximately 7\%, while "reversing" shows a performance gain of up to 60\%. This suggests that maneuver-based intentions strongly depend on extended historical context. Furthermore, the reduction in normalized Levenshtein distance indicates that increasing the length of observed trajectory improves sequence-level accuracy.

Future research should investigate augmenting the feature vector beyond velocity-based inputs, incorporating visual cues, and balancing the dataset to improve performance for underrepresented classes, e.g., "braking" and "reversing."

\section{Conclusion and Future Work}
\label{sec:conclusion}

This paper introduced the Emirates Multi-Task (EMT) dataset, designed to support the evaluation of tracking, trajectory prediction, and intention prediction within a unified multi-task framework. On cross-task evaluation, in detection-tracking pipeline, the fine-tuned detector significantly improves performance, reinforcing the importance of using region-specific data for effective adaptation of deep learning algorithms in autonomous vehicle applications. For trajectory prediction, we evaluate both sequential and interaction-aware models, providing training pipelines for sequential and frame-based learning. For intention prediction, two evaluation settings are provided: cross-validation and train/test splits. We report cross-validation results as a benchmark, while the train/test setting is intended for assessing model generalization on previously unseen intention classes. The broader cross-task investigations, as tracking-to-prediction, detection-to-tracking-to-prediction, tracking-to-intention inference, and joint models where trajectory forecasting is conditioned on predicted intentions are left open for future investigation by the research community and datasets users.

The evaluated algorithms operate solely on relative distances computed from absolute positions extracted from images. We leave it to the community to experiment with and design models that integrate visual cues for tracking and prediction. Additionally, the EMT dataset includes numerous highway scenarios that reflect regional traffic conditions. Balancing the dataset to address underrepresented patterns and improving model generalizability, considering the diversity gap between training and testing datasets, is left for researchers to explore.

The current dataset represents the first version collected from the region using a frontal camera. The primary direction for future work is to integrate Sim2Real scene generation to expand the dataset by including underrepresented scenarios. Additionally, we plan to collect a multimodal dataset incorporating LiDAR, camera data, and localization information. This dataset will address underrepresented scenarios identified during the analysis of the EMT dataset, ensuring a more comprehensive resource for the safe deployment of autonomous vehicles in the Gulf region. The dataset will also be sufficiently large to support large model training. For evaluation, future work will involve cross-evaluation by training models on multiple existing datasets and evaluating them on regional data. This process will incrementally include fine-tuning samples to assess generalization and model performance on rare scenarios.

\bibliographystyle{ieeetr}
\bibliography{refs.bib}

\begin{thebibliography}{10}

\bibitem{10.1109/ITSC.2018.8569552}
R.~Krajewski, J.~Bock, L.~Kloeker, and L.~Eckstein, ``The highd dataset: A drone dataset of naturalistic vehicle trajectories on german highways for validation of highly automated driving systems,'' in {\em 2018 21st International Conference on Intelligent Transportation Systems (ITSC)}, p.~2118–2125, IEEE Press, 2018.

\bibitem{Kotseruba2016JointAI}
I.~Kotseruba, A.~Rasouli, and J.~K. Tsotsos, ``Joint attention in autonomous driving (jaad),'' {\em ArXiv}, vol.~abs/1609.04741, 2016.

\bibitem{9008118}
A.~Rasouli, I.~Kotseruba, T.~Kunic, and J.~Tsotsos, ``Pie: A large-scale dataset and models for pedestrian intention estimation and trajectory prediction,'' in {\em 2019 IEEE/CVF International Conference on Computer Vision (ICCV)}, pp.~6261--6270, 2019.

\bibitem{9013045}
B.~Liu, E.~Adeli, Z.~Cao, K.-H. Lee, A.~Shenoi, A.~Gaidon, and J.~C. Niebles, ``Spatiotemporal relationship reasoning for pedestrian intent prediction,'' {\em IEEE Robotics and Automation Letters}, vol.~5, no.~2, pp.~3485--3492, 2020.

\bibitem{10.5555/3600270.3602131}
C.~Xu, W.~Ding, W.~Lyu, Z.~Liu, S.~Wang, Y.~He, H.~Hu, D.~Zhao, and B.~Li, ``Safebench: a benchmarking platform for safety evaluation of autonomous vehicles,'' in {\em Proceedings of the 36th International Conference on Neural Information Processing Systems}, NIPS '22, (Red Hook, NY, USA), Curran Associates Inc., 2022.

\bibitem{10.1145/3576841.3585930}
R.~S. Hallyburton, S.~Zhang, and M.~Pajic, ``Avstack: An open-source, reconfigurable platform for autonomous vehicle development,'' in {\em Proceedings of the ACM/IEEE 14th International Conference on Cyber-Physical Systems (with CPS-IoT Week 2023)}, ICCPS '23, (New York, NY, USA), p.~209–220, Association for Computing Machinery, 2023.

\bibitem{Geiger2012AreWR}
A.~Geiger, P.~Lenz, and R.~Urtasun, ``Are we ready for autonomous driving? the kitti vision benchmark suite,'' {\em 2012 IEEE Conference on Computer Vision and Pattern Recognition}, pp.~3354--3361, 2012.

\bibitem{nuscenes2019}
H.~Caesar, V.~Bankiti, A.~H. Lang, S.~Vora, V.~E. Liong, Q.~Xu, A.~Krishnan, Y.~Pan, G.~Baldan, and O.~Beijbom, ``{ nuScenes: A Multimodal Dataset for Autonomous Driving },'' in {\em 2020 IEEE/CVF Conference on Computer Vision and Pattern Recognition (CVPR)}, (Los Alamitos, CA, USA), pp.~11618--11628, IEEE Computer Society, June 2020.

\bibitem{9709630}
S.~Ettinger, S.~Cheng, B.~Caine, C.~Liu, H.~Zhao, S.~Pradhan, Y.~Chai, B.~Sapp, C.~Qi, Y.~Zhou, Z.~Yang, A.~Chouard, P.~Sun, J.~Ngiam, V.~Vasudevan, A.~McCauley, J.~Shlens, and D.~Anguelov, ``Large scale interactive motion forecasting for autonomous driving : The waymo open motion dataset,'' in {\em 2021 IEEE/CVF International Conference on Computer Vision (ICCV)}, pp.~9690--9699, 2021.

\bibitem{xue2019blvdbuildinglargescale5d}
J.~Xue, J.~Fang, T.~Li, B.~Zhang, P.~Zhang, Z.~Ye, and J.~Dou, ``Blvd: Building a large-scale 5d semantics benchmark for autonomous driving,'' 2019.

\bibitem{9710155}
H.~Girase, H.~Gang, S.~Malla, J.~Li, A.~Kanehara, K.~Mangalam, and C.~Choi, ``Loki: Long term and key intentions for trajectory prediction,'' in {\em 2021 IEEE/CVF International Conference on Computer Vision (ICCV)}, pp.~9783--9792, 2021.

\bibitem{Houston2020OneTA}
J.~L. Houston, G.~C.~A. Zuidhof, L.~Bergamini, Y.~Ye, A.~Jain, S.~Omari, V.~I. Iglovikov, and P.~Ondruska, ``One thousand and one hours: Self-driving motion prediction dataset,'' in {\em Conference on Robot Learning}, 2020.

\bibitem{Argoverse}
M.-F. Chang, J.~W. Lambert, P.~Sangkloy, J.~Singh, S.~Bak, A.~Hartnett, D.~Wang, P.~Carr, S.~Lucey, D.~Ramanan, and J.~Hays, ``Argoverse: 3d tracking and forecasting with rich maps,'' in {\em Conference on Computer Vision and Pattern Recognition (CVPR)}, 2019.

\bibitem{Argoverse2}
B.~Wilson, W.~Qi, T.~Agarwal, J.~Lambert, J.~Singh, S.~Khandelwal, B.~Pan, R.~Kumar, A.~Hartnett, J.~K. Pontes, D.~Ramanan, P.~Carr, and J.~Hays, ``Argoverse 2: Next generation datasets for self-driving perception and forecasting,'' in {\em Proceedings of the Neural Information Processing Systems Track on Datasets and Benchmarks (NeurIPS Datasets and Benchmarks 2021)}, 2021.

\bibitem{TrustButVerify}
J.~Lambert and J.~Hays, ``Trust, but verify: Cross-modality fusion for hd map change detection,'' in {\em Proceedings of the Neural Information Processing Systems Track on Datasets and Benchmarks (NeurIPS Datasets and Benchmarks 2021)}, 2021.

\bibitem{Geiger2013IJRR}
A.~Geiger, P.~Lenz, C.~Stiller, and R.~Urtasun, ``Vision meets robotics: The kitti dataset,'' {\em International Journal of Robotics Research (IJRR)}, 2013.

\bibitem{Liao2022PAMI}
Y.~Liao, J.~Xie, and A.~Geiger, ``{KITTI}-360: A novel dataset and benchmarks for urban scene understanding in 2d and 3d,'' {\em Pattern Analysis and Machine Intelligence (PAMI)}, 2022.

\bibitem{10.1609/aaai.v33i01.33016120}
Y.~Ma, X.~Zhu, S.~Zhang, R.~Yang, W.~Wang, and D.~Manocha, ``Trafficpredict: Trajectory prediction for heterogeneous traffic-agents,'' AAAI'19/IAAI'19/EAAI'19, AAAI Press, 2019.

\bibitem{9156329}
F.~Yu, H.~Chen, X.~Wang, W.~Xian, Y.~Chen, F.~Liu, V.~Madhavan, and T.~Darrell, ``Bdd100k: A diverse driving dataset for heterogeneous multitask learning,'' in {\em 2020 IEEE/CVF Conference on Computer Vision and Pattern Recognition (CVPR)}, pp.~2633--2642, 2020.

\bibitem{zhan2019interactiondatasetinternationaladversarial}
W.~Zhan, L.~Sun, D.~Wang, H.~Shi, A.~Clausse, M.~Naumann, J.~Kummerle, H.~Konigshof, C.~Stiller, A.~de~La~Fortelle, and M.~Tomizuka, ``Interaction dataset: An international, adversarial and cooperative motion dataset in interactive driving scenarios with semantic maps,'' 2019.

\bibitem{8296962}
N.~Wojke, A.~Bewley, and D.~Paulus, ``Simple online and realtime tracking with a deep association metric,'' in {\em 2017 IEEE International Conference on Image Processing (ICIP)}, pp.~3645--3649, 2017.

\bibitem{9010033}
P.~Bergmann, T.~Meinhardt, and L.~Leal-Taixé, ``Tracking without bells and whistles,'' in {\em 2019 IEEE/CVF International Conference on Computer Vision (ICCV)}, pp.~941--951, 2019.

\bibitem{zhang2021fairmot}
Y.~Zhang, C.~Wang, X.~Wang, W.~Zeng, and W.~Liu, ``Fairmot: On the fairness of detection and re-identification in multiple object tracking,'' {\em International Journal of Computer Vision}, vol.~129, pp.~3069--3087, 2021.

\bibitem{10.1007/978-3-030-58621-8_7}
Z.~Wang, L.~Zheng, Y.~Liu, Y.~Li, and S.~Wang, ``Towards real-time multi-object tracking,'' in {\em Computer Vision – ECCV 2020: 16th European Conference, Glasgow, UK, August 23–28, 2020, Proceedings, Part XI}, (Berlin, Heidelberg), p.~107–122, Springer-Verlag, 2020.

\bibitem{BoT-SORT}
N.~Aharon, R.~Orfaig, and B.-Z. Bobrovsky, ``Bot-sort: Robust associations multi-pedestrian tracking,'' {\em arXiv preprint arXiv:2206.14651}, 2022.

\bibitem{bytetrack}
Y.~Zhang, P.~Sun, Y.~Jiang, D.~Yu, F.~Weng, Z.~Yuan, P.~Luo, W.~Liu, and X.~Wang, ``Bytetrack: Multi-object tracking by associating every detection box,'' 2022.

\bibitem{10406854}
N.~A. Madjid, A.~Sharma, B.~Hassan, N.~Werghi, J.~Dias, and M.~Khonji, ``Multi-target tracker for low light vision,'' in {\em 2023 21st International Conference on Advanced Robotics (ICAR)}, pp.~252--257, 2023.

\bibitem{10160328}
M.~Nagy, M.~Khonji, J.~Dias, and S.~Javed, ``Dfr-fastmot: Detection failure resistant tracker for fast multi-object tracking based on sensor fusion,'' in {\em 2023 IEEE International Conference on Robotics and Automation (ICRA)}, pp.~827--833, 2023.

\bibitem{cao2023observation}
J.~Cao, J.~Pang, X.~Weng, R.~Khirodkar, and K.~Kitani, ``Observation-centric sort: Rethinking sort for robust multi-object tracking,'' in {\em Proceedings of the IEEE/CVF Conference on Computer Vision and Pattern Recognition}, pp.~9686--9696, 2023.

\bibitem{7780479}
A.~Alahi, K.~Goel, V.~Ramanathan, A.~Robicquet, L.~Fei-Fei, and S.~Savarese, ``Social lstm: Human trajectory prediction in crowded spaces,'' in {\em 2016 IEEE Conference on Computer Vision and Pattern Recognition (CVPR)}, pp.~961--971, 2016.

\bibitem{Varshneya2017HumanTP}
D.~Varshneya and G.~Srinivasaraghavan, ``Human trajectory prediction using spatially aware deep attention models,'' {\em ArXiv}, vol.~abs/1705.09436, 2017.

\bibitem{zhang2019srlstmstaterefinementlstm}
P.~Zhang, W.~Ouyang, P.~Zhang, J.~Xue, and N.~Zheng, ``Sr-lstm: State refinement for lstm towards pedestrian trajectory prediction,'' in {\em 2019 IEEE/CVF Conference on Computer Vision and Pattern Recognition (CVPR)}, (Los Alamitos, CA, USA), pp.~12077--12086, IEEE Computer Society, June 2019.

\bibitem{Manh2018SceneLSTMAM}
H.~T. Manh and G.~Alaghband, ``Scene-lstm: A model for human trajectory prediction,'' {\em ArXiv}, vol.~abs/1808.04018, 2018.

\bibitem{Hasan2018MXLSTMMT}
I.~Hasan, F.~Setti, T.~Tsesmelis, A.~D. Bue, F.~Galasso, and M.~Cristani, ``Mx-lstm: Mixing tracklets and vislets to jointly forecast trajectories and head poses,'' {\em 2018 IEEE/CVF Conference on Computer Vision and Pattern Recognition}, pp.~6067--6076, 2018.

\bibitem{Xu2023UncoveringTM}
Y.~Xu, A.~Bazarjani, H.~gun Chi, C.~Choi, and Y.~R. Fu, ``Uncovering the missing pattern: Unified framework towards trajectory imputation and prediction,'' {\em 2023 IEEE/CVF Conference on Computer Vision and Pattern Recognition (CVPR)}, pp.~9632--9643, 2023.

\bibitem{8954462}
R.~Chandra, U.~Bhattacharya, A.~Bera, and D.~Manocha, ``Traphic: Trajectory prediction in dense and heterogeneous traffic using weighted interactions,'' in {\em 2019 IEEE/CVF Conference on Computer Vision and Pattern Recognition (CVPR)}, pp.~8475--8484, 2019.

\bibitem{10.1109/IVS.2018.8500493}
N.~Deo and M.~M. Trivedi, ``Multi-modal trajectory prediction of surrounding vehicles with maneuver based lstms,'' in {\em 2018 IEEE Intelligent Vehicles Symposium (IV)}, p.~1179–1184, IEEE Press, 2018.

\bibitem{Zyner2018NaturalisticDI}
A.~Zyner, S.~Worrall, and E.~M. Nebot, ``Naturalistic driver intention and path prediction using recurrent neural networks,'' {\em IEEE Transactions on Intelligent Transportation Systems}, vol.~21, pp.~1584--1594, 2018.

\bibitem{zhang_lstm_2023}
C.~Zhang, Z.~Ni, and C.~Berger, ``Spatial-temporal-spectral lstm: A transferable model for pedestrian trajectory prediction,'' {\em IEEE Transactions on Intelligent Vehicles}, vol.~PP, pp.~1--14, 01 2023.

\bibitem{8317913}
F.~Altché and A.~de~La~Fortelle, ``An lstm network for highway trajectory prediction,'' in {\em 2017 IEEE 20th International Conference on Intelligent Transportation Systems (ITSC)}, pp.~353--359, 2017.

\bibitem{yu2020spatiotemporalgraphtransformernetworks}
C.~Yu, X.~Ma, J.~Ren, H.~Zhao, and S.~Yi, ``Spatio-temporal graph transformer networks for pedestrian trajectory prediction,'' in {\em Computer Vision – ECCV 2020: 16th European Conference, Glasgow, UK, August 23–28, 2020, Proceedings, Part XII}, (Berlin, Heidelberg), p.~507–523, Springer-Verlag, 2020.

\bibitem{Zhou2022GASTTHT}
L.~Zhou, D.~Yang, X.~Zhai, S.~Wu, Z.~Hu, and J.~Liu, ``Ga-stt: Human trajectory prediction with group aware spatial-temporal transformer,'' {\em IEEE Robotics and Automation Letters}, vol.~7, pp.~7660--7667, 2022.

\bibitem{9878832}
Z.~Zhou, L.~Ye, J.~Wang, K.~Wu, and K.~Lu, ``Hivt: Hierarchical vector transformer for multi-agent motion prediction,'' in {\em 2022 IEEE/CVF Conference on Computer Vision and Pattern Recognition (CVPR)}, pp.~8813--8823, 2022.

\bibitem{9710708}
Y.~Yuan, X.~Weng, Y.~Ou, and K.~Kitani, ``Agentformer: Agent-aware transformers for socio-temporal multi-agent forecasting,'' in {\em 2021 IEEE/CVF International Conference on Computer Vision (ICCV)}, pp.~9793--9803, 2021.

\bibitem{Amirloo2022LatentFormerMT}
E.~Amirloo, A.~Rasouli, P.~Lakner, M.~Rohani, and J.~Luo, ``Latentformer: Multi-agent transformer-based interaction modeling and trajectory prediction,'' {\em ArXiv}, vol.~abs/2203.01880, 2022.

\bibitem{Li2022GraphbasedST}
L.~Li, M.~Pagnucco, and Y.~Song, ``Graph-based spatial transformer with memory replay for multi-future pedestrian trajectory prediction,'' {\em 2022 IEEE/CVF Conference on Computer Vision and Pattern Recognition (CVPR)}, pp.~2221--2231, 2022.

\bibitem{zhang_2022}
K.~Zhang, X.~Feng, L.~Wu, and Z.~He, ``Trajectory prediction for autonomous driving using spatial-temporal graph attention transformer,'' {\em IEEE Transactions on Intelligent Transportation Systems}, vol.~PP, pp.~1--11, 11 2022.

\bibitem{10504962}
Y.~Liu, B.~Li, X.~Wang, C.~Sammut, and L.~Yao, ``Attention-aware social graph transformer networks for stochastic trajectory prediction,'' {\em IEEE Transactions on Knowledge \& Data Engineering}, vol.~36, pp.~5633--5646, Nov. 2024.

\bibitem{9577819}
Y.~Liu, J.~Zhang, L.~Fang, Q.~Jiang, and B.~Zhou, ``Multimodal motion prediction with stacked transformers,'' in {\em 2021 IEEE/CVF Conference on Computer Vision and Pattern Recognition (CVPR)}, pp.~7573--7582, 2021.

\bibitem{Feng_2023}
C.~Feng, H.~Zhou, H.~Lin, Z.~Zhang, Z.~Xu, C.~Zhang, B.~Zhou, and S.~Shen, ``Macformer: Map-agent coupled transformer for real-time and robust trajectory prediction,'' {\em IEEE Robotics and Automation Letters}, vol.~8, p.~6795–6802, Oct. 2023.

\bibitem{9812226}
Z.~Su, G.~Huang, S.~Zhang, and W.~Hua, ``Crossmodal transformer based generative framework for pedestrian trajectory prediction,'' in {\em 2022 International Conference on Robotics and Automation (ICRA)}, pp.~2337--2343, 2022.

\bibitem{8917228}
X.~Li, X.~Ying, and M.~C. Chuah, ``Grip: Graph-based interaction-aware trajectory prediction,'' in {\em 2019 IEEE Intelligent Transportation Systems Conference (ITSC)}, pp.~3960--3966, 2019.

\bibitem{Li2019GRIPEG}
X.~Li, X.~Ying, and M.~C. Chuah, ``Grip++: Enhanced graph-based interaction-aware trajectory prediction for autonomous driving,'' {\em arXiv: Computer Vision and Pattern Recognition}, 2019.

\bibitem{liang2020learning}
M.~Liang, B.~Yang, R.~Hu, Y.~Chen, R.~Liao, S.~Feng, and R.~Urtasun, ``Learning lane graph representations for motion forecasting,'' in {\em ECCV}, 2020.

\bibitem{10.1007/978-3-030-58523-5_40}
T.~Salzmann, B.~Ivanovic, P.~Chakravarty, and M.~Pavone, ``Trajectron++: Dynamically-feasible trajectory forecasting with heterogeneous data,'' in {\em Computer Vision – ECCV 2020: 16th European Conference, Glasgow, UK, August 23–28, 2020, Proceedings, Part XVIII}, (Berlin, Heidelberg), p.~683–700, Springer-Verlag, 2020.

\bibitem{e294141389194a54a05536938fcdd509}
V.~Kosaraju, A.~Sadeghian, R.~Mart\'{\i}n-Mart\'{\i}n, I.~Reid, S.~H. Rezatofighi, and S.~Savarese, {\em Social-BiGAT: multimodal trajectory forecasting using bicycle-GAN and graph attention networks}.
\newblock Red Hook, NY, USA: Curran Associates Inc., 2019.

\bibitem{CHENG2023163}
H.~Cheng, M.~Liu, L.~Chen, H.~Broszio, M.~Sester, and M.~Y. Yang, ``Gatraj: A graph- and attention-based multi-agent trajectory prediction model,'' {\em ISPRS Journal of Photogrammetry and Remote Sensing}, vol.~205, pp.~163--175, 2023.

\bibitem{Wang2020GraphTCNSI}
C.~Wang, S.~Cai, and G.~S.~H. Tan, ``Graphtcn: Spatio-temporal interaction modeling for human trajectory prediction,'' {\em 2021 IEEE Winter Conference on Applications of Computer Vision (WACV)}, pp.~3449--3458, 2020.

\bibitem{Liao2024MFTrajMB}
H.~Liao, Z.~Li, C.~Wang, H.~Shen, B.~Wang, D.~Liao, G.~Li, and C.~Xu, ``Mftraj: Map-free, behavior-driven trajectory prediction for autonomous driving,'' in {\em International Joint Conference on Artificial Intelligence}, 2024.

\bibitem{10.1109/ICRA46639.2022.9812253}
T.~Gilles, S.~Sabatini, D.~Tsishkou, B.~Stanciulescu, and F.~Moutarde, ``Gohome: Graph-oriented heatmap output for future motion estimation,'' in {\em 2022 International Conference on Robotics and Automation (ICRA)}, p.~9107–9114, IEEE Press, 2022.

\bibitem{Westny2023MTPGOGP}
T.~Westny, J.~Oskarsson, B.~Olofsson, and E.~Frisk, ``Mtp-go: Graph-based probabilistic multi-agent trajectory prediction with neural odes,'' {\em IEEE Transactions on Intelligent Vehicles}, vol.~8, pp.~4223--4236, 2023.

\bibitem{10205349}
C.~Xu, R.~T. Tan, Y.~Tan, S.~Chen, Y.~G. Wang, X.~Wang, and Y.~Wang, ``{ EqMotion: Equivariant Multi-Agent Motion Prediction with Invariant Interaction Reasoning },'' in {\em 2023 IEEE/CVF Conference on Computer Vision and Pattern Recognition (CVPR)}, (Los Alamitos, CA, USA), pp.~1410--1420, IEEE Computer Society, June 2023.

\bibitem{7487409}
V.~Karasev, A.~Ayvaci, B.~Heisele, and S.~Soatto, ``Intent-aware long-term prediction of pedestrian motion,'' in {\em 2016 IEEE International Conference on Robotics and Automation (ICRA)}, pp.~2543--2549, 2016.

\bibitem{Deo2018HowWS}
N.~Deo, A.~Rangesh, and M.~M. Trivedi, ``How would surround vehicles move? a unified framework for maneuver classification and motion prediction,'' {\em IEEE Transactions on Intelligent Vehicles}, vol.~3, pp.~129--140, 2018.

\bibitem{Kim2017ProbabilisticVT}
B.~Kim, C.~M. Kang, J.~Kim, S.-H. Lee, C.~C. Chung, and J.~W. Choi, ``Probabilistic vehicle trajectory prediction over occupancy grid map via recurrent neural network,'' {\em 2017 IEEE 20th International Conference on Intelligent Transportation Systems (ITSC)}, pp.~399--404, 2017.

\bibitem{10.1109/IVS.2017.7995919}
A.~Zyner, S.~Worrall, J.~Ward, and E.~Nebot, ``Long short term memory for driver intent prediction,'' in {\em 2017 IEEE Intelligent Vehicles Symposium (IV)}, p.~1484–1489, IEEE Press, 2017.

\bibitem{Phillips2017GeneralizableIP}
D.~J. Phillips, T.~A. Wheeler, and M.~J. Kochenderfer, ``Generalizable intention prediction of human drivers at intersections,'' {\em 2017 IEEE Intelligent Vehicles Symposium (IV)}, pp.~1665--1670, 2017.

\bibitem{9712346}
G.~Singh, S.~Akrigg, M.~Maio, V.~Fontana, R.~Alitappeh, S.~Khan, S.~Saha, K.~Jeddisaravi, F.~Yousefi, J.~Culley, T.~Nicholson, J.~Omokeowa, S.~Grazioso, A.~Bradley, G.~Gironimo, and F.~Cuzzolin, ``Road: The road event awareness dataset for autonomous driving,'' {\em IEEE Transactions on Pattern Analysis \& Machine Intelligence}, vol.~45, pp.~1036--1054, jan 2023.

\bibitem{Murad}
M.~Mebrahtu, A.~Araia, A.~Ghebreslasie, J.~Dias, and M.~Khonji, ``Transformer-based multi-modal probabilistic pedestrian prediction for risk-aware autonomous vehicle navigation,'' in {\em 2023 21st International Conference on Advanced Robotics (ICAR)}, pp.~652--659, 2023.

\bibitem{4160958}
L.~Yujian and L.~Bo, ``A normalized levenshtein distance metric,'' {\em IEEE Transactions on Pattern Analysis and Machine Intelligence}, vol.~29, no.~6, pp.~1091--1095, 2007.

\end{thebibliography}

%\newpage
%\onecolumn

%\appendices
%\section{Annotation Labels}

%\input{tables/agents_table}
%\input{tables/action_table}

%\newpage 
%\section{Sample Frames and Annotations}
%\begin{figure*}[h!]
%\centering
%\includegraphics[width=0.9\textwidth]{figures/appendix_1.pdf}
%\caption{Sample annotated frames (Day)}
%\label{fig:samples_day}
%\end{figure*}
%\begin{figure*}[h!]
%\centering
%\includegraphics[width=0.9\textwidth]{figures/appendix_2.pdf}
%\caption{Sample annotated frames (Night)}
%\label{fig:samples_night}
%\end{figure*}

%\section{}

%\section*{Acknowledgment}


\ifCLASSOPTIONcaptionsoff
  \newpage
\fi

\end{document}